\documentclass[lettersize,journal, onecolumn]{IEEEtran}
\usepackage{amsmath,amsfonts}
\usepackage{algorithmic}
\usepackage{algorithm}
\usepackage{array}
\usepackage[caption=false,font=normalsize,labelfont=sf,textfont=sf]{subfig}
\usepackage{textcomp}
\usepackage{stfloats}
\usepackage{url}
\usepackage{verbatim}
\usepackage{graphicx}
\usepackage{cite}
\begin{document}

\title{Stochastic Hessian Fittings with Lie Groups}

\author{
Xi-Lin Li, lixilinx@gmail.com
}



\maketitle

\begin{abstract}
This report investigates the fitting of the Hessian or its inverse for stochastic optimizations using a Hessian fitting criterion derived from the preconditioned stochastic gradient descent (PSGD) method. This criterion is closely related to many widely used second-order and adaptive gradient optimization methods, including BFGS, the Gauss-Newton algorithm, natural gradient descent, and AdaGrad. Our analyses reveal the efficiency and reliability differences of a broad range of preconditioner fitting methods, ranging from closed-form to iterative approaches, using Hessian-vector products or stochastic gradients only, with Hessian fittings across various geometric settings (the Euclidean space, the manifold of symmetric positive definite (SPD) matrices, and a variety of Lie groups). The most intriguing finding is that the Hessian fitting problem is strongly convex under mild conditions in certain general Lie groups. This result turns Hessian fitting into a well-behaved Lie group optimization problem and facilitates the design of highly efficient and elegant Lie group sparse preconditioner fitting methods for large-scale stochastic optimizations. 
\end{abstract}

\begin{IEEEkeywords}
Stochastic optimization, stochastic gradient descent (SGD), Hessian-vector product, Lie group, strong convexity.
\end{IEEEkeywords}

\section{Introduction}
This report considers the problem of Hessian fitting for second-order stochastic optimizations where the objective functions are typically defined as an expectation, say $
f(\theta) =  E_z[\ell(\theta, z)] 
$ with $\theta\in \mathbb{R}^n$ being the parameters to be optimized, and $z$ a random variable or vector that can be sampled to evaluate an unbiased estimate of $f(\theta)$. 
We generally assume that $f(\theta)$ is second-order differentiable such that a preconditioner $P\succ 0$, say the inverse of the Hessian of $f(\theta)$ when it is convex, can be used to accelerate the convergence of the following stochastic gradient descent (SGD) iteration,
\begin{equation}\label{psgd}
\theta_{t+1} = \theta_{t} - \mu_t P_t \,{\partial f_t(\theta)}/{\partial \theta}\left.\right|_{\theta=\theta_t}
\end{equation}
to an optimal solution for $\theta$, where the subscript $t$ is the time index, and $f_t(\theta)$ is a sampled noisy estimation of $f(\theta)$ at step $t$. To simplify the notation, let us ignore the time index $t$ when doing so does not cause misunderstanding, and use $g={\partial f(\theta)}/{\partial \theta}$ and $h={\partial ( v^T g)}/{\partial \theta}=Hv$ to denote the stochastic gradient and Hessian-vector product associated with a vector $v\sim \mathcal{N}(0,I)$ at a certain iteration step, respectively, where $H={\partial^2 f(\theta)}/{(\partial \theta^T \partial \theta)}$ is the Hessian matrix, and we typically do not assume its definiteness. Then, with the following preconditioner estimation criterion proposed in \cite{Li18},
\begin{equation}\label{P_criterion}
    c(P; v, h) = h^T P h + v^T P^{-1} v
\end{equation}
we can derive most of the commonly used preconditioners as listed in Table I \cite{psgd_affine} by solving for $P$ from $\partial{c}/\partial{P}=0$. We will show that the criterion (\ref{P_criterion}) is convex such that $\partial{c}/\partial{P}=0$  determines a unique solution for $P$. 
\begin{enumerate}
\item When the curvature condition $v^Th>0$ is met, the solution $Phh^TP=vv^T$ reduces to the secant equation $Ph=v$, which is used to derive the classic BFGS and limited-memory BFGS (L-BFGS) like quasi-Newton methods via iterative Hessian fittings \cite{Boyd}.  These off-the-shelf methods are not popular for stochastic optimizations as they do not take the gradient noises into consideration. 
\item Closed-form solution $P=(H^2)^{-0.5}$ reduces to Newton's method when $H\succ 0$, i.e., $P=H^{-1}$. The diagonal form of this solution is particularly popular in machine learning due to its simplicity, for example, ESGD \cite{ESGD}, AdaHessian \cite{AdaHessian}, and Sophia \cite{Sophia} to name a few. 
\item When the target is to minimize a negative log-likelihood (NLL) loss and $g_z={\partial \ell(\theta, z)}/{\partial \theta}$ is a per-sample gradient for each $z$, closed-form solution $P^{-2}= E_z[g_zg_z^T] $ reduces to the (empirical) Fisher information matrix, whose inverse is used in the Gauss-Newton and natural gradient methods \cite{gn_kfac, natural_gradient} as preconditioners. Its Kronecker factorized approximate solution is popular for learning matrix parameters \cite{KFAC, hess_kron, gn_kfac}.     
\item Closed-form solution $P_t=\left(\sum_{i=1}^t g_ig_i^T\right)^{-0.5}$ reduces to the preconditioner used in the AdaGrad \cite{adagrad}, a large family of optimizers that includes many popular examples, for example, the RMSProp \cite{rmsprop} and Adam(W)  \cite{Adam} that use diagonal preconditioners, and a few methods with the name Shampoo the most well known \cite{hess_kron, Shampoo1,Shampoo2,Shampoo3} that use Kronecker factorized approximate closed-form solutions.   
\end{enumerate}

Except for the trivial case of diagonal preconditioners, all the last three closed-form solutions for $P$ in Table I require matrix or matrix square root inverse operations, which could cause numerical difficulties with finite precision arithmetic \cite{Shampoo2}. Furthermore, exact closed-form solutions are rarely available for sparse Hessian estimations. The iterative Hessian fitting method of BFGS requires that $v^Th$ is well above zero, which is hardly met in stochastic optimizations. The preconditioned stochastic gradient descent (PSGD) method fits $P$ iteratively in a variety of Lie groups \cite{Li18, psgd_affine, li2022black}, and has successfully avoided the numerical difficulties of these closed-form solutions and also lifted the limitations of BFGS. Recent empirical results have further confirmed the excellent performance of PSGD in a handful of large scale machine learning related stochastic optimizations \cite{asdl, pooladzandi2024curvatureinformed}.  Still, there lacks a deeper theoretical understanding of the pros and cons of the preconditioners listed in Table I, disregarding their certain practical limitations, such as computational complexities and numerical difficulties with finite precision arithmetic. One major contribution of this report is to show that the problem of Hessian fitting by minimizing criterion (\ref{P_criterion}) converges linearly to the optimal solution in the manifold of symmetric positive definite (SPD) matrices, and furthermore, is strongly convex in certain Lie groups, see Proposition 5 in Section III.A, while those closed-form solutions only converge sublinearly. Another significant contribution is the invention of a L-BFGS like low-rank approximation (LRA) version of PSGD \footnote{This LRA preconditioner fitting work has been presented in the NeurIPS HOOML workshop, 2022 \cite{li2022black}.}. Lastly, the insights brought up by the studies here also help to improve previous PSGD implementations \cite{Li18, psgd_affine, li2022black}  in several aspects, including better step size control for preconditioner fitting and four inverse-free versions of PSGD.     

The notations in this paper are fairly standard. For example, $A\succ 0$ means that $A$ is a SPD matrix, $E_v[\cdot]$ takes expectation with respect to a certain random variable $v$, $\mathcal{N}(0,I)$ denotes the standard normal distribution, $A^T$ is the transpose of $A$, $A^{0.5}$ denotes the principal square root of  $A$, $A^{-2T}$ is the shorthand for $(A^{-2})^T$, $\|A\|_2$ or simply $\|A\|$ is the spectral norm of $A$, ${\rm tr}(A)$ takes the trace of $A$, $\lambda_{\min}(A)$ takes the minimum eigenvalue of $A$, ${\rm vec}(A)$ vectorizes matrix $A$ by stacking its columns successively, GL$(n, \mathbb{R})$ stands for the general linear group of $n\times n$ invertible matrices, and $\otimes$ denotes the Kronecker product. We say that $A$ and $B$ commute if $[A,B]=AB-BA=0$. For a symmetric matrix $A\in\mathbb{R}^{n\times n}$, we only need to vectorize its upper or lower triangular part, and denote this operation as ${\rm vech}(A)$. Clearly, ${\rm vec}(A)=S\,{\rm vech}(A)$, where $S\in\mathbb{R}^{n^2\times [0.5n(n+1)]}$ is a sparse full column rank matrix with elements being $0$ or $1$. By convention, we usually use capital letters for matrices, and lowercase letters for vectors and scalars. 

\begin{table}[h]
\caption{Variations of criterion (\ref{P_criterion}) for preconditioner fitting, where $v$ is a dummy variable in 2), 3) and 4) as $E_v[v^TP^{-1}v]={\rm tr}(P^{-1})$.   }
\begin{center}
\begin{tabular}{ |c|c| } 
 \hline
 Criterion & Solution for $P$   \\ 
 \hline 
 1) $c(P;v,h)$ & $Phh^TP=vv^T$   \\ 
 2) $E_v[c(P;v,h)]$ & $P^{-2}=H^2 $   \\
 3) $E_{v,z}[c(P;v,g_z)]$ & $P^{-2}= E_z[g_zg_z^T] $   \\
 4) $ \sum_{i=1}^t E_{v_i}\left[ c(P;v_i,g_i)\right]$ & $P_t^{-2}= \sum_{i=1}^t g_ig_i^T $  \\
 \hline
\end{tabular}
\end{center}
\end{table}

\section{Candidates for Stochastic Hessian Fittings: an Overview}

\subsection{The Problem Setups}

We are to fit the Hessian or its inverse sequentially, i.e., updating the Hessian estimate once given every pair of vector and its associated Hessian-vector product, $(v, h)$, with model,
\begin{equation}\label{hv_model}
    h = Hv + \epsilon
\end{equation}
where $H\in \mathbb{R}^{n\times n}$ is the unknown Hessian matrix to be estimated, and $\epsilon$ denotes any possible modeling errors, e.g., truncation and quantization errors when the Hessian-vector product $h$ is approximated with a finite-difference method, or the randomness error when a stochastic yet unbiased proxy of the target function $f(\theta) =  E_z[\ell(\theta, z)]$  is adopted for Hessian-vector product evaluation, or the structural errors caused by using a certain form of sparse matrices to fit dense Hessians. We assume that $\epsilon$ is a random error term independent of $v$.   

We choose to update $P$ by minimizing the preconditioner fitting criterion  (\ref{P_criterion}), whose expectation with respect to $v$ is,
\begin{align}\nonumber
    c(P) = & E_{v\sim\mathcal{N}(0,I)}[c(P; v, h)] \\ \nonumber
     = & {\rm tr} \left( P E[ hh^T] + P^{-1}E[vv^T] \right) \\ \label{cP}
     = & {\rm tr}\left( P\left( H^2 + E[\epsilon\epsilon^T] \right) + P^{-1} \right)
\end{align}
Then, the optimal SPD solution minimizing $c(P)$ can be shown to be \cite{Li18}
\begin{equation}\label{opt_P_e}
    P=\left(H^2 + E[\epsilon\epsilon^T] \right)^{-0.5}
\end{equation} 
by solving equation ${\partial c(P)}/{\partial P}=0$ for $P$. One can also replace the $P$ in (\ref{P_criterion}) with $P^{-1}$ to obtain another criterion for fitting the Hessian directly.  Here, we only consider the fitting criterion (\ref{P_criterion}) since the intention is to use $P$ as a preconditioner to accelerate the SGD iteration in (\ref{psgd}). For the gradient whitening preconditioner, we simply replace the pair $(v, h)$ in (\ref{hv_model}) with $(v, g)$, where $v$ is a dummy or auxiliary variable and can be optionally integrated out as shown in (\ref{cP}). 

As revealed by (\ref{opt_P_e}), the error term $\epsilon$ in (\ref{hv_model}) helps damp the preconditioner estimation and has been shown to regularize the optimizer in a perfectly balanced way \cite{Li18}. The direct use of $(H^{2})^{-0.5}$ as a preconditioner could severely amplify the gradient noises, leading to divergence \cite{Li18}. The definiteness of Hessian $H$ is important for the optimization of $f(\theta)$, but not relevant to our setups, since both (\ref{cP}) and (\ref{opt_P_e}) suggest that the signs of $\lambda(H)$ do not matter to the finial solutions for $P$. Hence, without loss of generality, we always assume that $H$ is a constant positive definite matrix in our performance studies. Please note that, except for the BFGS method, other preconditioners derived from (\ref{P_criterion}) are always valid regardless of the definiteness of $H$.   

We will further ignore the term $\epsilon$ most of the time to simplify our performance analyzes. Still, note that $c(P)$ always has form $c(P)={\rm tr}(PA+P^{-1})$, where $A=H^2 + E[\epsilon\epsilon^T]$ with pair $(v,h)$, and $A=E_z[g_zg_z^T]$ or $\sum_{i=1}^t [g_ig_i^T]$ with pair $(v, g)$.  Thus, the properties proved for one case can be easily transferred to the other cases.  

The criterion (\ref{P_criterion}) is a well defined function for any $P$ with $\det(P)\ne 0$, i.e., the  group GL$(n, \mathbb{R})$. For our purposes, we further require $P\succ 0$ so that it is a valid preconditioner for the SGD iteration in (\ref{psgd}). The implicitly implied symmetry constraint, i.e., $P=P^T$, will slightly complicate the calculations of the derivatives of $c(P;v,h)$ wrt (with respect to) $P$. We will explicitly mention this subtleness when it matters. From now on, by default, the gradient and Hessian always refer to those of $c(P;v,h)$ or $c(P)=E_v[c(P;v,h)]$ wrt $P$. Gradient descent (GD) or SGD also refers only to the iterations for updating $P$, not $\theta$, if not further clarified.       

\subsection{Second Order Approximation of Criterion (\ref{P_criterion}) and Its Convexity}

Let $dP$ denote a small perturbation around $P$. We are to expand the preconditioner fitting loss in (\ref{P_criterion}) up to its second order terms of $dP$. To begin with, we expand $(P+dP)^{-1}$ around $P^{-1}$ as,
\begin{align}\nonumber
    & \quad dP^{-1}  =  (P+dP)^{-1} - P^{-1} \\ \nonumber
    &  = (I - P^{-1} dP + P^{-1} dP\, P^{-1} dP + \mathcal{O}((dP)^3) )P^{-1} - P^{-1} \\ \label{d_inv_P}
   &   =  -P^{-1} dP\, P^{-1} + P^{-1} dP\, P^{-1} dP \, P^{-1} + \mathcal{O}((dP)^3)
\end{align}
where $\mathcal{O}((dP)^3)$ includes the third and higher order terms of $dP$ that are negligible when $\|dP\|$ is small enough. Now, with (\ref{d_inv_P}), we can expand (\ref{P_criterion}) as below,
\begin{align}\nonumber
     dc(P; v, h) =  & h^T dP\, h + v^T dP^{-1}\, v \\ \label{dcP_2nd}
     =  & h^T dP\, h  - v^T P^{-1} dP\, P^{-1} v  +  v^T P^{-1} dP\, P^{-1} dP \, P^{-1} v 
\end{align}
where we have suppressed the term $\mathcal{O}((dP)^3)$ to simplify our equation. 
Clearly, the $dc(P; v, h)$ in (\ref{dcP_2nd}) becomes a quadratic function of $dP$ after truncating its higher order terms.  
To maximize the reduction of $c(P;v,h)$, we can solve for the optimal $dP$ by letting $d c(P; v, h) /d P=0$, i.e., the following  equation, 
\begin{align}\label{opt_dP}
	&  hh^T - P^{-T} v v^T P^{-T} + P^{-T}vv^TP^{-T} dP^T\, P^{-T}    + P^{-T} dP^T\, P^{-T}vv^T P^{-T} = 0
\end{align}
where we do not consider the symmetry constraint $P=P^T$ yet to simplify the calculation of derivatives. 
From (\ref{opt_dP}), it is ready to identify the gradient of $c(P;v,h)$ with respect to  $P $ as below,
\begin{equation}\label{cP_gradient}
    \frac{\partial c(P; v, h)}{\partial P} = hh^T - {P}^{-T} v v^T {P}^{-T}
\end{equation}
Note that (\ref{cP_gradient}) gives the steepest ascent direction of $c(P;v,h)$ in the space $\mathbb{R}^{n\times n}$. With the symmetry constraint $P=P^T$, we should double the off-diagonal elements of the gradient in (\ref{cP_gradient}) to obtain the steepest ascent direction in the space $\mathbb{R}^{0.5n(n+1)}$. We can choose either form for the update of $P$ with GD or SGD. However, we need to calculate the Hessian with respect to $P$ in the space $\mathbb{R}^{0.5n(n+1)}$. We typically take the expectation of criterion (\ref{P_criterion}), i.e., the $c(P) = E_{v\sim \mathcal{N}(0,I)}[c(P;v, h)]$ given in (\ref{cP}), for the performance analysis. There is the following convex property of $c(P)$.

\emph{Proposition 1: }Criterion $c(P)$ is a convex function of $P$ in the set $\{P|P\succ 0\}$. 

\emph{Proof: }From (\ref{dcP_2nd}), we can rewrite the second order term of $dc(P)$ as 
\begin{align}\nonumber 
    & \quad E_{v\sim\mathcal{N}(0,I)}\left[ v^T P^{-1} dP\, P^{-1} dP \, P^{-1} v  \right] \\ \nonumber 
    & = {\rm tr}\left(E_v\left[   P^{-1} dP\, P^{-1} dP \, P^{-1} v v^T  \right] \right) \\ \nonumber 
    & = {\rm tr}\left(   P^{-1} dP\, P^{-1} dP \, P^{-1}  \right) \\ \nonumber 
    & = \left({\rm vec}\left( \left( P^{-1} dP \right)^T \right)\right)^T {\rm vec}\left( P^{-1} dP\, P^{-1} \right) \\ \nonumber 
    & = \left({\rm vec}\left( dP \, P^{-1} \right)\right)^T {\rm vec}\left( P^{-1} dP\, P^{-1} \right) \\ \nonumber 
    & = \left( \left(P^{-1}\otimes I\right) {\rm vec}(dP) \right)^T \left( \left(P^{-1} \otimes P^{-1} \right) {\rm vec}(dP) \right) \\ \nonumber 
    & = \left(  {\rm vec}(dP) \right)^T \left(P^{-1}\otimes I\right) \left(P^{-1} \otimes P^{-1} \right) {\rm vec}(dP) \\ \nonumber 
    & = \left(  {\rm vec}(dP) \right)^T \left(P^{-2}\otimes P^{-1}\right)   {\rm vec}(dP)  \\ \label{proof_proposition1}
    & = \left(  {\rm vech}(dP) \right)^T S^T \left(P^{-2}\otimes P^{-1}\right)   S\,{\rm vech}(dP)
\end{align}
where we have used the relation ${\rm tr}(A^TB) $$=$$ \left({\rm vec}(A)\right)^T {\rm vec}(B)$ to arrive at the fourth line from the third one, ${\rm vec}(ABC) = (C^T\otimes A){\rm vec}(B)$ from the fifth  to the sixth line,  $(AB)\otimes (CD) = (A\otimes C)(B\otimes D)$  from the third to the last to the next line, and definition ${\rm vec}(A)=S\,{\rm vech}(A)$ for $A\succ 0$ to get the last line. Since $P^{-2}\otimes P^{-1} \succ 0$ and $S$ has full column rank, the Hessian with respect to $P$, i.e., 
$$S^T \left(P^{-2}\otimes P^{-1}\right)   S$$
due to the last line of (\ref{proof_proposition1}), is positive definite. 
\hfill $\square$

Note that following the proof of Proposition 1, we can also show that $c(P;v,h)$ is convex as well. In fact, this convexity property is independent of the definiteness of $H$. In general, the quadratic term in (\ref{dcP_2nd}) can be negative without the implicit constraint $dP=dP^T$ even when $P\succ 0$. Thus, the symmetry constraint is necessary for the convexity of $c(P)$. 

\subsection{Hessian Fitting in the Euclidean Space, the Manifold of SPD Matrices, and Lie Groups}

By Proposition 1, it is expected that our Hessian fitting problem is always convex in these spaces. However, it is not always strongly convex. Specifically, it is not strongly convex in the Euclidean space and the manifold of SPD matrices. Thus, it could be difficult to develop efficient Hessian fitting methods there. We put the detailed discussions on Hessian fitting in those two types of spaces in Appendix I since they are not the focus of this report. On the other hand, we can show that the Hessian fitting problem can be strongly convex in certain Lie groups. Furthermore, we can tightly lower bound the Hessian of $c(P)$ there, which facilitates the development of Hessian fitting methods in Lie groups. We leave those details to the next section.   

We summarize most of the above Hessian fitting candidates in Table II. To confirm our analysis, Fig.~1 shows a set of typical convergence curves when $H$ is a small Hilbert matrix. Except for Newton method, which starts from $P_0=0.02I$, all the other methods start from $P_0=I$. The step sizes for SGD in the Euclidean space and the manifold of SPD matrices are tuned to achieve the maximum rate of convergence. Step sizes for SGD with Lie group and Newton's method are already normalized to $1$. Except for the SGD (\ref{cP_hv_gd}) in the Euclidean space and the closed-form solution (\ref{closed-form-solutionI}) that converge sublinearly, all the other methods converge to $H^{-1}$ within a reasonable number of iterations. These empirical convergence rates are well in line with the predictions in Table II. 

\begin{table}[h]
\caption{Candidates for Hessian fitting. Note that $0<\rho<1$ is not necessarily the same for different methods. }
\begin{center}
\begin{tabular}{ |c|c|c|c| } 
 \hline
 Method & Equation & Space & Error at $t$ \\ 
 \hline 
 SGD & (\ref{cP_hv_gd}) & Euclidean & $\mathcal{O}(1/\sqrt{t})$ \\ 
 Closed-form & (\ref{closed-form-solutionI}) & Euclidean & $\mathcal{O}(1/t)$ \\
 SGD & (\ref{GD_on_SPD_manifold}) & SPD & $\mathcal{O}(\rho^t)$ \\
 SGD & (\ref{standard_method}) & Lie group & $\mathcal{O}(\rho^t)$ \\
 Newton & (\ref{newton_P}) & Euclidean/SPD & $\mathcal{O}(\rho^{2^t})$ \\
 \hline
\end{tabular}
\end{center}
\end{table}

\begin{figure}[h]
\centering
\includegraphics[width=0.6\textwidth]{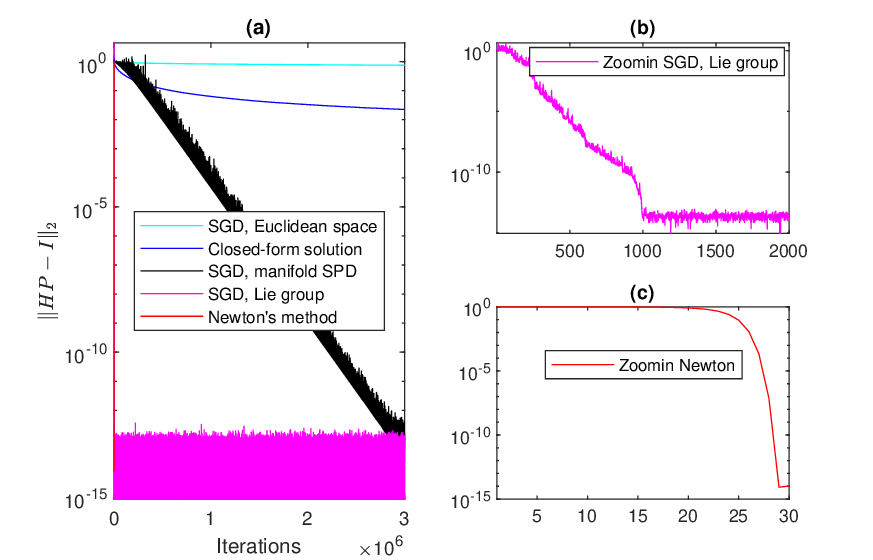}
\caption{(a): Typical convergence curves for the five Hessian fitting methods in Table II when  $H$ is a $3\times 3$ Hilbert matrix, i.e., $H_{i,j}=1/(i+j-1)$ with $1\le i,j\le n$. (b) and (c):  Zooming in of the converge curves of SGD in group GL$(n, \mathbb{R})$ and Newton's method, respectively, for better visualization.    }
\end{figure}

\section{Hessian Fittings in GL$(n, \mathbb{R})$}

\subsection{Strong Convexity in GL$(n, \mathbb{R})$}

We let
\begin{equation}\label{p_qtq}
    P = Q^T Q
\end{equation}
where the factor $Q$ is in a certain Lie group. When $Q$ is a triangular matrix, (\ref{p_qtq}) is known as the Cholesky decomposition or factorization of $P$. However, in general, $Q$ could take many possible forms. It is natural to come up with (\ref{p_qtq}) since with a $\theta$-dependent coordinate change $\vartheta=Q^{-T}\theta$, the preconditioned SGD update in (\ref{psgd}) becomes a standard SGD in the new coordinate system $\vartheta$. Lie group provides a natural tool for seeking such a coordinate transform. The group structure also helps to preserve certain invariances for transform $\theta\rightarrow\vartheta$, e.g., orientation with the group GL$^+(n,\mathbb{R})$, vector length with the group O$(n)$, etc. Such geometrical constraints actually help to regularize the update of $Q$ in a rigid way. In this section, we focus mainly on the Hessian fitting in GL$(n, \mathbb{R})$. 

With the Lie group constraint, we let $I + \mathcal{E}=e^{\varepsilon G}$, where $\varepsilon\rightarrow 0$ is a small scalar such that $\varepsilon \|G\|\ll 1$, $G$ is the group's infinitesimal generator, and $\mathcal{E}=\varepsilon G + \varepsilon^2 G^2/2! + \ldots$ is a small matrix. Then, we can have the following two choices for $dQ$, 
\begin{align}\nonumber 
    {\rm choice \,I:}\quad & dQ =  e^{\varepsilon G} Q - Q = \mathcal{E} Q \\ \label{dQ_two_choices}
    {\rm choice \,II:}\quad & dQ =  Qe^{\varepsilon G} - Q   =  Q\mathcal{E} 
\end{align}
Correspondingly, the $dP$ in (\ref{p_qtq}) will have the following two possible forms, 
\begin{align}\nonumber 
    {\rm choice \,I:}\quad  dP = & (Q+\mathcal{E}Q)^T(Q+\mathcal{E}Q) - P \\ \nonumber
    = & Q^T(\mathcal{E} + \mathcal{E}^T + \mathcal{E}^T\mathcal{E}) Q \\ \nonumber 
    {\rm choice \,II:}\quad  dP = & (Q+Q\mathcal{E})^T(Q+Q\mathcal{E}) - P \\ \label{dP_two_choices}
    = & \mathcal{E}^TP + P\mathcal{E} + \mathcal{E}^T P \mathcal{E}
\end{align}
The second choice of $dP$ in (\ref{dP_two_choices}) looks almost the same as that in (\ref{local_coordinate_spd}) after ignoring the second order term of $\mathcal{E}$. But now $\mathcal{E}$ is a small matrix associated with the group generator and is no longer required to be symmetric. Substituting (\ref{dP_two_choices}) back into (\ref{dcP_2nd}) and noting that $P^{-1}=Q^{-1}Q^{-T}$, we can arrive at the following two possible choices for $dc(P; v, h)$,
\begin{align}\nonumber 
    &  dc_{I}(P;v,h) =  h^TQ^T(\mathcal{E}+ \mathcal{E}^T) Qh - v^TQ^{-1}  (\mathcal{E}+ \mathcal{E}^T) Q^{-T} v   + h^TQ^T \mathcal{E}^T \mathcal{E} Qh + v^TQ^{-1} (\mathcal{E}^2 + \mathcal{E}\mathcal{E}^T  +  \mathcal{E}^{2T}) Q^{-T} v  \\ \label{dcP_two_choices}
    &   dc_{II}(P;v,h) =  h^T(\mathcal{E}^TP + P\mathcal{E})h  - v^T(P^{-1}\mathcal{E}^T + \mathcal{E}P^{-1})v  + h^T\mathcal{E}^T P \mathcal{E}h + v^T(\mathcal{E}^2P^{-1} + \mathcal{E}P^{-1}\mathcal{E}^T + P^{-1}\mathcal{E}^{2T})v 
\end{align}
where we have suppressed higher order terms $\mathcal{O}(\mathcal{E}^3)$ to shorten our equations. 
The first choice in (\ref{dcP_two_choices}) not only looks more balanced and symmetric, but also has nicer properties than the second one as will be revealed in our following discussions. Thus, we mainly focus on the first choice.   

One striking difference between Hessian fitting in the Lie groups and the manifold of SPD matrices is that now criterion $c(P)$ could be strongly convex under certain mild conditions such as $\lambda_{\min}(H) > 0$. Clearly, $c(P)$ cannot be strictly convex in GL$(n, \mathbb{R})$ in the conventional sense, since $Q$ and $UQ$ give the same value for $c(P)$ with any orthogonal matrix $U$. But $c(P)$ could behave much better after properly separating out this rotation ambiguity. To this end, we need to turn to the polar decomposition \cite{Golub} $A=U\Gamma$ to define a proper manifold that is independent of such rotations, where $A\in \mathbb{R}^{n\times n}$ is a square matrix, $U$ is an orthogonal matrix, and $\Gamma\succeq 0$ is semi-positive definite. With the polar decomposition, let us introduce the following equivalence relation. 

\emph{Equivalence relation $R_{\rm polar}$}: For $Q_1$ and $Q_2$ in $\mathbb{R}^{n\times n}$, we say that $Q_1\sim Q_2$ if and only if their polar decompositions share the same $\Gamma$, i.e., $Q_1=U_1\Gamma$ and $Q_2=U_2\Gamma$ for certain orthogonal matrices $U_1$ and $U_2$.  

It is not difficult to show that $R_{\rm polar}$ indeed defines an equivalence relation by demonstrating that it is reflexive, symmetric, and transitive.  
Then, we can define the quotient set GL$(n, \mathbb{R})/R_{\rm polar}$.  Note that since the Lie group $O(n)$ has dimension $0.5n(n-1)$, elements from this quotient set have dimension $n^2 - 0.5n(n-1)=0.5n(n+1)$. Indeed, its elements can be represented as SPD matrices after discarding the rotation ambiguity. Hence, the SPD manifold can be defined as a quotient manifold in this way. 

\emph{Proposition 5: }Criterion $c(P)$ is strongly convex in the quotient set GL$(n, \mathbb{R})/R_{\rm polar}$ with the local coordinate system induced by choice I in (\ref{dQ_two_choices}) when $\lambda_{\min}(H)\ge \lambda_0 >0$.

\emph{Proof: }First, let us briefly show that with $A\succ 0$ and eigenvalue decomposition $A=UDU^T$, $aA+bA^{-1}- 2\sqrt{ab} I =U(aD+bD^{-1} - 2\sqrt{ab})U^T\succeq 0$, where $a$ and $b$ are two positive numbers. 

Second, we show that it is sufficient to consider a symmetric $\mathcal{E}$ in the quotient set GL$(n, \mathbb{R})/R_{\rm polar}$. Starting from $Q$, let $Q_1 = Q + \mathcal{E}_1Q$ be an arbitrary neighbor of $Q$ in the quotient set GL$(n, \mathbb{R})/R_{\rm polar}$ with   $\mathcal{E}_1$ small enough such that $\|\mathcal{E}_1\|_2 < 1$. The condition, $\|\mathcal{E}_1\|_2 < 1$, is to ensure that $Q_1$ is invertible.   Let the polar decomposition of $Q_1$ and $I+\mathcal{E}_1$ be
\[Q_1 = U_1 \Gamma_1, \qquad I+\mathcal{E}_1 = U_2 \Gamma_2 \]
respectively. Then, we have, 
\[\Gamma_2 Q =  U_2^T(I + \mathcal{E}_1) Q 
    = U_2^T Q_1 =
     (U_2^TU_1) \Gamma_1 \]
Thus, by letting $\mathcal{E}_2 = \Gamma_2 - I$, we have, 
\begin{equation}\label{Q1-Q2}
    (I+ \mathcal{E}_2)Q =(U_2^TU_1) \Gamma_1\sim U_1 \Gamma_1 = (I+ \mathcal{E}_1)Q
\end{equation}
Since $\Gamma_2$ is symmetric, $\mathcal{E}_2 = \Gamma_2 - I$ is symmetric as well. Thus, (\ref{Q1-Q2}) suggests that it is enough to consider only a symmetric $\mathcal{E}$ in the quotient set GL$(n, \mathbb{R})/R_{\rm polar}$. This is not surprising, as a symmetric $\mathcal{E}$ just has dimension $0.5n(n+1)$.  
 
The last step is to check the lower bound of the second order terms of $dc_I(P)$ in (\ref{dcP_two_choices}). Now, without loss of generality, let us assume that $\mathcal{E}=\mathcal{E}^T$ due to (\ref{Q1-Q2}). Starting from (\ref{dcP_two_choices}), a lower bound of the expectation of the second order terms of $dc_I(P;v,h)$ with choice $dQ = \mathcal{E}Q$ is given by
\begin{align*}
    & E_v\left[ h^TQ^T \mathcal{E}^T \mathcal{E} Qh + v^TQ^{-1} (\mathcal{E}^2 + \mathcal{E}\mathcal{E}^T  +  \mathcal{E}^{2T}) Q^{-T} v \right] \\
   = &  {\rm tr}\left( \mathcal{E}^2 Q H^2 Q^T + 3\mathcal{E}^2 Q^{-T} Q^{-1} \right) \\
   \ge & {\rm tr}\left( \left(\lambda_{\min}(H)\right)^2 \mathcal{E}^2  Q Q^T + 3\mathcal{E}^2 \left(QQ^T\right)^{-1} \right) \\
   \ge & {\rm tr}\left( \mathcal{E}^2\left( \lambda_0^2 QQ^T + 3(QQ^T)^{-1} \right) \right) \\
   \ge & 2\sqrt{3} \lambda_0 \, {\rm tr}\left( \mathcal{E}^2\right) \\
   = & 2\sqrt{3} \lambda_0 \left({\rm vec}(\mathcal{E})\right)^T {\rm vec}(\mathcal{E}) \\
   = & 2\sqrt{3} \lambda_0 \left({\rm vech}(\mathcal{E})\right)^T S^TS \, {\rm vech}(\mathcal{E}) \\
   = &  \left({\rm vech}(\mathcal{E})\right)^T \left( 2\sqrt{3} \lambda_0 \, S^TS\right) {\rm vech}(\mathcal{E})
\end{align*}
where we have used expectations $E_v[hh^T]=H^2$ and $E_v[vv^T]=I$ from the first to the second line, $aA+bA^{-1} \succ 2\sqrt{ab} I$ for $A\succ 0$ to arrive the fifth line from the fourth one, and definition ${\rm vec}(A)=S{\rm vech}(A)$ for a symmetric matrix $A$ to arrive the last line. Thus, the Hessian of $c_I(P)$  in the new coordinate is lower bounded by $2\sqrt{3} \lambda_0 \, S^TS \succ 2\sqrt{3} \lambda_0 I$ since $S^TS\succ I$. 
\hfill $\square$ 

We briefly present a few more statements regarding the convexity of the Hessian fitting problem in Lie groups without formal proofs. Clearly, $c(P)$ is strongly convex in any connected group of diagonal matrices when $\lambda_{\min}(H)\ge \lambda_0 >0$.  With group $O(n)$, from (\ref{dcP_two_choices}),  it is not difficult to show that every point is a saddle point since $\mathcal{E}$$=-\mathcal{E}^T$.  The second choice, $c_{II}(P)$, is generally not convex, even with the constraint $\mathcal{E}=\mathcal{E}^T$. Empirical results have also confirmed that Hessian fitting with choice II converges significantly slower than that with choice I. The first choice, $c_I(P)$, is generally not convex in the group of triangular matrices, although previous results have shown that Hessian fitting with this group does work well \cite{Li18}.  An explanation is provided in the next subsection. Lastly, since we do not care about the rotation ambiguity for preconditioning purposes, Proposition 5 suggests that $c(P)$ is ``virtually'' strongly convex in the group GL$(n, \mathbb{R})$ with the local coordinate system induced by choice I in (\ref{dQ_two_choices}) when $\lambda_{\min}(H)\ge \lambda_0 >0$. 

\subsection{The Two  Basic Lie Group Hessian Fitting Methods }

\subsubsection{Fitting in  Group GL$(n, \mathbb{R})$}
 
As our Hessian fitting criterion is ``virtually'' strongly convex in this group, we can easily follow the conventional convex optimization theory to design proper Hessian fitting methods \cite{Boyd}. If we could find an upper bound for the eigenvalues of the Hessian of $c(P)$, say $L$, we know that GD or SGD with step size $1/L$ converges linearly to the optimal solution $H^{-1}$ at least with rate $1-2\sqrt{3}\lambda_0/L$, where $2\sqrt{3}\lambda_0$ is the lower bound of the eigenvalues of the Hessian of $c(P)$ derived from the proof of Proposition 5. From (\ref{dcP_two_choices}), choice I, we know that the sum of the second order terms of $\mathcal{E}$ is upper bounded by
\begin{align}\nonumber 
  & \left| h^TQ^T \mathcal{E}^T \mathcal{E} Qh + v^TQ^{-1} (\mathcal{E}^2 + \mathcal{E}\mathcal{E}^T  +  \mathcal{E}^{2T}) Q^{-T} v \right| \\ \nonumber 
  \le & \left| h^TQ^T \mathcal{E}^T \mathcal{E} Qh \right| + \left| v^TQ^{-1} \mathcal{E}^2  Q^{-T} v \right|  + \left| v^TQ^{-1} \mathcal{E}\mathcal{E}^T  Q^{-T} v \right| + \left| v^TQ^{-1} \mathcal{E}^{2T}  Q^{-T} v \right| \\ \label{L_rough_est}
     \le & \left( \|Qh\|^2  + 3   \|Q^{-T}v\|^2   \right) \|\mathcal{E}\|^2
\end{align}
where $\|\mathcal{E}\|$ can be any matrix norm for $\mathcal{E}$. 
Thus, from (\ref{L_rough_est}), we see that a rough estimate of $L$ could be 
\begin{equation}\nonumber 
    L_t\approx \max_{1\le i\le t}\left( \|Q_ih_i\|^2 + 3\|Q_i^{-T}v_i\|^2\right) 
\end{equation}
The above bound is too pessimistic. In our implementations, we take the following more relaxed one, 
\begin{equation}\label{Lt}
    \ell_t = \|Q_th_t\|^2 + \|Q_t^{-T}v_t\|^2, \quad L_t= \max\left(\beta L_{t-1} + (1-\beta)\ell_t, \; \ell_t\right) 
\end{equation}
where $0\le \beta\le 1$.  A large $\beta$ is preferred when the pair $(v, h)$ has a sparse distribution over time.  
At the same time, from the linear terms in (\ref{dcP_two_choices}), choice I, we see that the gradient in group GL$(n, \mathbb{R})$  is 
\begin{equation}\label{gradient_glg}
    \nabla_{Q'} = 2\left( Qhh^TQ^T - Q^{-T}vv^TQ^{-1}\right)
\end{equation}  
which also is the gradient in the quotient set GL$(n, \mathbb{R})/R_{\rm polar}$ since it is already symmetric, where $Q'=\int (dQ) Q^{-1}$ gives the new coordinate. 
Combining (\ref{Lt}) and (\ref{gradient_glg}) gives the following first baseline Lie group SGD Hessian fitting method,
\begin{equation}\label{standard_method}
    Q_{t+1} = Q_t - \frac{\mu}{L_t} (Q_th_th_t^TQ_t^T - Q_t^{-T}v_tv_t^TQ_t^{-1} )   Q_t
\end{equation}
where $0<\mu\le 2$ is a normalized step size. 

A note on the matrix inverse in (\ref{standard_method}). Straightforward implementation of (\ref{standard_method}) needs to solve the linear system $Q^Tx=v$ to calculate $Q^{-T}v$. Still, from (\ref{standard_method}), we see that $Q_{t+1}-Q_t$ is a matrix of rank two. Thus, we can recursively update the inverse of $Q$ with the Woodbury matrix identity \cite{Golub} as done with the BFGS method. One concern is that the numerical errors may accumulate after repeatedly applying the Woodbury matrix identity many times.  Empirical results have confirmed that this procedure is generally numerically stable with double and single precision arithmetic, but could be unstable with half precision arithmetic. 

\subsubsection{Fitting in the Group of Triangular Matrices} 

We can simply take the triangular part of (\ref{gradient_glg}) to get the gradient for updating $Q$ in the group of triangular matrices as we have done in \cite{Li18}. However, in general $c(P)$ is not convex in this group. To take the advantage of Proposition 5, we can factorize out the rotation part of the updated $Q$ from (\ref{standard_method}) to have the following update formula for $Q$ in the group of triangular matrices,  
\begin{equation}\label{triangular_update} 
    Q_{t+1} = \left[I - \frac{\mu}{L_t} (Q_th_th_t^TQ_t^T - Q_t^{-T}v_tv_t^TQ_t^{-1} ) \right]_R  Q_t
\end{equation}
where $[A]_R$ only keeps the triangular part of the QR decomposition of a square matrix $A$. In this way, $Q_t$ is still in the quotient set GL$(n, \mathbb{R})/R_{\rm polar}$, but keeps to be a triangular matrix as long as $Q_0$ is. Note that only two rank-1 QR decomposition updates are required to obtain the QR decomposition of the matrix inside the operator $[\cdot]_R$ in (\ref{triangular_update}) due to its low-rank structure \cite{Golub}.  One pleasant by-product is that now we can calculate  $Q_t^{-T}v_t$  with backward substitution, which is cheap and empirically shown to be numerically stable even with half precision arithmetic. 

Let us derive an approximation for $[A]_R$ to reveal the connection between (\ref{triangular_update}) and the method in \cite{Li18}. Let $\|\Delta\|\ll 1$ be a small symmetric matrix, and consider the following QR decomposition,
\begin{align}\nonumber 
I+\Delta & = (I+\Delta_1)(I+\Delta_2) \\  \label{qr_approx1}
I & = (I+\Delta_1)^T(I+\Delta_1)    
\end{align}
where $I+\Delta_1$ is an orthogonal matrix and $\Delta_2$ is an upper triangular matrix such that the first equation of (\ref{qr_approx1}) gives the QR decomposition for $I+\Delta$. Note that $\Delta_1$ and $\Delta_2$ are small matrices as well since QR decomposition is numerically stable.  After expanding (\ref{qr_approx1}) and ignoring the second order terms $\Delta_1^T\Delta_1$ and $\Delta_1\Delta_2$, we can solve for $\Delta_1$ and $\Delta_2$ jointly from the following equations,
$$
\Delta_1 + \Delta_2  \approx \Delta, \qquad \Delta_1+\Delta_1^T  \approx 0
$$
which can be shown to be
\begin{align*}
\Delta_1 & \approx{\rm tril}(\Delta, -1) - {\rm triu}(\Delta, 1) \\
\Delta_2& \approx {\rm triu}(\Delta) + {\rm triu}(\Delta, 1)
\end{align*}
when $[\cdot]_R$ takes the upper triangular part, where ${\rm tril}(A, -1)$ and ${\rm triu}(A, 1)$ exclude the diagonal elements of ${\rm tril}(A)$ and ${\rm triu}(A)$, respectively. Hence, we obtain the following approximation for $[A]_
R$,
\begin{equation}\label{qr_approx2}
\left[I+\Delta \right]_R \approx I + {\rm triu}(\Delta) + {\rm triu}(\Delta, 1), \qquad {\rm for}\, \|\Delta\|\ll 1
\end{equation}
In this way, we see that the gradient in the group of triangular matrices still points to an ascent direction for updating $Q$ in the group GL$(n, \mathbb{R})$, although not the steepest one. It is safe to take the approximation (\ref{qr_approx2}) for the $[\cdot]_R$ operation in (\ref{triangular_update}) for small update step sizes, say $\mu\le 0.1$. Numerical results have confirmed that after separating out this rotation ambiguity, the preconditioner update method (\ref{triangular_update})  converges faster than the one given in \cite{Li18}. Yet, for the case of large step sizes, the approximation in (\ref{qr_approx2}) loses precision. Then, it is better to simply let $\left[I+\Delta \right]_R \approx I + {\rm triu}(\Delta)$ for $0\ll \|\Delta\|< 1$. 

\subsubsection{Newton's Methods} 

Lastly, we want to briefly mention Newton's methods for preconditioner fitting with Lie groups, which can be derived from (\ref{dcP_two_choices}) by solving for the optimal $\mathcal{E}$ that reduces $c(P)$ the most. But they turn out to be significantly more complicated than the form in (\ref{newton_P}), and seem to have little practical value. Nevertheless, the Hessian for the case considered in Proposition 5 still has the following neat form,
\[ S^T\left[ I\otimes (QH^2Q^T + 3 Q^{-T}Q^{-1})\right]S  \]

\subsection{Numerical Results}

Fig.~2 shows a few typical Hessian fitting convergence curves for the PSGD Hessian fitting methods with (\ref{standard_method}) and (\ref{triangular_update}) using $\beta=0$ and step size $\mu=1$, the closed-form solution (\ref{closed-form-solutionI}) by clipping the maximum exponential moving average factor to $0.999$, and the classic BFGS formula \cite{Boyd}.  We have tested the following three cases. 

\subsubsection{Clean Hessian-Vector Products}

In this case, we set the ground truth $H'$ to $H$. All methods except for the closed-form one can converge to the ground truth solution up to errors only bounded by the machine precision.  The PSGD style with (\ref{standard_method}) and BFGS have higher error plateaus than the PSGD style with (\ref{triangular_update}) due to the accumulated numerical errors caused by repeated usage of the Woodbury matrix identity \cite{Golub} for recursive matrix inverse updating. 

\subsubsection{Noisy Hessian-Vector Products}

We have tested a noisy model of (\ref{hv_model}) with $\epsilon \sim \mathcal{N}(0, \sigma_{\epsilon}^2I) $ and $\sigma_{\epsilon}=0.01$. Hence, we set the ground truth to $H'=(H^2+\sigma_{\epsilon}^2I)^{0.5}$ due to (\ref{opt_P_e}). We see that the BFGS is very sensitive to noises, and diverges easily.  The two PSGD methods with reduced step size $\mu=0.1$ converge slower than the closed-form solution initially, but reach lower fitting errors eventually. Note that now there always are certain steady state preconditioner fitting errors with PSGD since the gradient in (\ref{gradient_glg}) cannot settle down to zero when $\sigma_{\epsilon}>0$. Yet, the preconditioner fitting problem is still strongly convex. One can show this by following the proof of Proposition 5 and replacing the $H$ there with $(H^2+\sigma_{\epsilon}^2I)^{0.5}$. If necessary, the preconditioner fitting errors of PSGD can be sufficiently small  given enough iterations to anneal down the step sizes. 

\subsubsection{Time-Varying Hessians}

The Hessians are from a time-varying process defined by $H_{t+1}=H_t + uu^T$, where $u_i\sim \mathcal{U}(0,1)$ for $1\le i\le n$, and $H_0$ is a matrix with all elements being $1/4$. The ground truth $H'=H_t$ is also time-varying. The BFGS diverges first as the Hessians change too fast and only locks onto the target $H_t^{-1}$ after $t>1000$. The closed-form solution has difficulty in tracking a time-varying Hessian due to its sublinear convergence.  The two PSGD style Hessian fitting methods lock onto the target quicker than BFGS, and also do not show any sign of  divergence before convergence. 

\begin{figure}[h]
\centering
\includegraphics[width=0.7\textwidth]{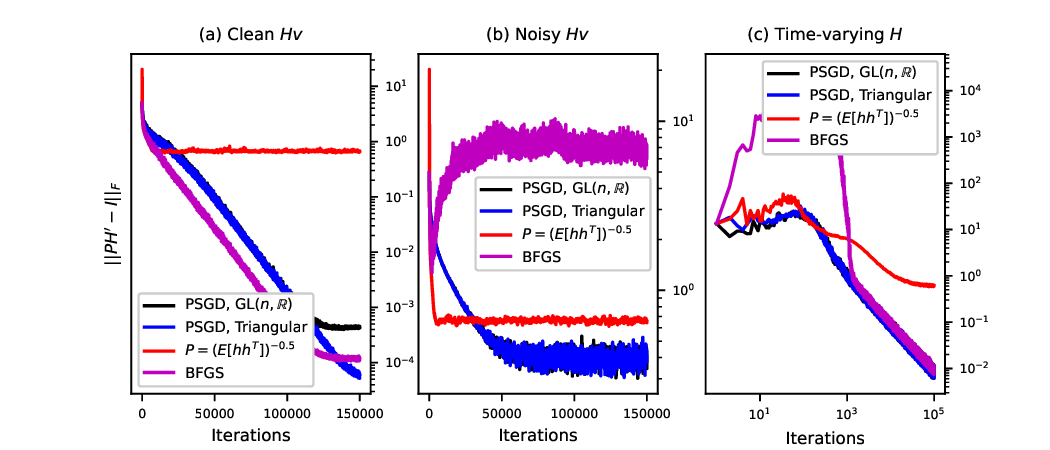}
\caption{(a) Typical convergence curves when $H$ is a $50\times 50$  matrix with $H_{i,i}=1$ and $H_{i,j}=0.5$ for $|i-j|=1$. Its eigenvalues are bounded in range $(0, 2)$. (b) The same $H$ as in (a), but with a noisy model of (\ref{hv_model}), where $\epsilon \sim \mathcal{N}(0, \sigma_{\epsilon}^2I )$, and $\sigma_{\epsilon}=0.01$. (c)  Time varying Hessians defined by process $H_{t+1}=H_t + uu^T$, where $u_i\sim \mathcal{U}(0,1)$ for $1\le i\le n=50$, and $H_0$ is a matrix with all elements being $1/4$.  }
\end{figure}

\section{Inverse-Free Hessian Fitting Methods}

One main drawback of the Hessian fitting methods in (\ref{standard_method}) and (\ref{triangular_update}) is that they either need to maintain the inverse of $Q$ or require a triangular solver to update the preconditioner. Here, we further develop a few inverse-free Hessian fitting methods that have the same multiplicative update form for $Q$ as $Q\leftarrow Q - \mu(aa^T - bb^T)Q$ or $Q\leftarrow Q - \mu Q(aa^T - bb^T)$, where $\|\mu(aa^T + bb^T)\|<1$ such that $Q$ always is in GL$(n, \mathbb{R})$.  

\subsection{Hessian Fitting with Local Coordinate $dQ=Q\mathcal{E}Q$}

We have mentioned in Section III that the local coordinate $dQ=Q\mathcal{E}$ does not lead to an efficient Hessian fitting method. Still, from (\ref{dcP_two_choices}), we see that the gradient wrt $Q'=\int Q^{-1}dQ$ is 
\[ \nabla_{Q'} = 2 (Phh^T - vv^TP^{-1}) \]
We can readily remove the term $P^{-1}$ in the above gradient by right-multiplying it with $P$. This gives the following inverse-free update rule for $Q$,
\begin{equation}\label{update_QEQ}
Q_{t+1} = Q_t - \frac{\mu}{L_t} Q_t (P_t h_th_t^T P_t - v_t v_t^T) 
\end{equation}
where similar to (\ref{Lt}), $L_t$ can be estimated as 
\begin{equation}\label{hess_bound_qeq} 
    \ell_t = \|P_th_t\|^2 + \|v_t\|^2, \quad L_t= \max\left(\beta L_{t-1} + (1-\beta)\ell_t, \; \ell_t\right) 
\end{equation}
It is not difficult to show that with the new coordinate $dQ=Q\mathcal{E}Q$, i.e., $Q'=\int Q^{-1}(dQ)Q^{-1}$, the gradient wrt to $Q'$ and the second order terms of $\mathcal{E}$ for the preconditioner fitting criterion (\ref{P_criterion}) are 
\begin{align*}
    & \nabla_{Q'} = 2(Phh^TQ^T - vv^T Q^{-1}) \\
    & h^T Q^T \mathcal{E}^T P\mathcal{E}Qh  + v^TQ^{-1}\mathcal{E}^TQ^T\mathcal{E}^Tv + v^T \mathcal{E}Q\mathcal{E}Q^{-T}v + v^T\mathcal{E}\mathcal{E}^T v
\end{align*}
respectively. Hence, (\ref{update_QEQ}) actually gives the gradient descent method for $Q$ minimizing criterion (\ref{P_criterion}) with the local coordinate $dQ=Q\mathcal{E}Q$. 
Similar to the case with local coordinate $dQ=Q\mathcal{E}$, generally, criterion (\ref{P_criterion}) is not necessarily convex in this new coordinate. Furthermore, the above quadratic terms do not provide us with a clean way to bound the Hessian wrt to $Q'$. The bound given in (\ref{hess_bound_qeq}) can be significantly underestimated for ill-conditioned $Q$, resulting in oscillating convergence curves for fitting ill-conditioned $H$. Nevertheless, we have found that (\ref{update_QEQ}) works well in practice. 

\subsection{Hessian Fitting with Local Coordinate $dQ=Q^{0.5} \mathcal{E}Q^{1.5}$}

By Proposition 5, we can assume that $Q$ is SPD. Then, we can precondition the gradient in (\ref{gradient_glg}) by both left- and right-multiplying it with $Q$ to remove the two matrix inverses. Considering that the local coordinate $dQ=\mathcal{E}Q$ is used to derive the gradient in (\ref{gradient_glg}), we can obtain the following inverse-free update rule for $Q$,
\begin{equation}\label{update_quad1}
Q_{t+1} = Q_t - \frac{\mu}{L_t} (P_t h_th_t^T P_t - v_t v_t^T) Q_t  
\end{equation}
where $L_t$ is the same as in (\ref{update_QEQ}). Again, we can show that with the new local coordinate $dQ=Q^{0.5} \mathcal{E}Q^{1.5}$, the gradient wrt $Q'=\int Q^{-0.5} (dQ) Q^{-1.5}$ and the second order terms of $\mathcal{E}$ for criterion (\ref{P_criterion}) are 
\begin{align*}
    & \nabla_{Q'} = 2(Q^{1.5}hh^T Q^{1.5} - Q^{-0.5}vv^TQ^{-0.5}) \\
    & h^T Q^{1.5} \mathcal{E}^T Q \mathcal{E} Q^{1.5}h + v^T Q^{-0.5} \mathcal{E}Q\mathcal{E} Q^{-0.5}v + v^T Q^{-0.5} \mathcal{E}Q\mathcal{E}^T  Q^{-0.5}v + v^T Q^{-0.5} \mathcal{E}^TQ\mathcal{E}^T Q^{-0.5}v
\end{align*}
respectively. 
Thus, (\ref{update_quad1}) gives the gradient descent method for $Q$ minimizing criterion (\ref{P_criterion}) in this new local coordinate. It is also not difficult to show that the Hessian fitting loss (\ref{P_criterion}) is strongly convex there with constraint $\mathcal{E}=\mathcal{E}^T$. However, (\ref{update_quad1}) alone cannot provide a practical Hessian fitting method, since it requires condition $Q\succ 0$ to define the local coordinate  $dQ=Q^{0.5} \mathcal{E}Q^{1.5}$. We need to make $Q$ SPD again with transform $Q\leftarrow (Q^TQ)^{0.5}$ whenever it is not. Calculating the matrix square root reliably and accurately can be expensive, especially with low precision arithmetic. Here, we provide two practical methods for fitting $Q$ in this geometry.

\subsubsection{Online Rotation Imposing the SPD Property}

Clearly, $Q$ will lose the SPD property after being repetitively updated with (\ref{update_quad1}). One way to pull it back to the set of SPD matrices is to find an orthogonal matrix $\Omega$ to rotate $Q$ as $Q\leftarrow \Omega Q$, where $\Omega$ can be solved from the following orthogonal Procrustes problem,
\begin{equation}\label{opb}
     \min_{\Omega} \|\Omega Q - I \|_F^2, \quad {\rm s.t.} \; \Omega^T\Omega = I
\end{equation}
Then, $\Omega Q$ produces the SPD matrix in the polar decomposition of $Q$. It is not difficult to show that (\ref{opb}) is equivalent to the following problem,
\begin{equation}\label{max_tr}
    \max_{\Omega} {\rm tr}(\Omega Q), \quad {\rm s.t.} \; \Omega^T\Omega = I
\end{equation}
Such a rotation matrix can be generated as $\Omega = e^{aR}$, where $R$ is a skew-symmetric matrix and $a$ is the rotation step size. The optimal $R'=aR$ can be determined by solving $\partial {\rm tr}(e^{R'}Q)/\partial R' = 0$ after expanding $e^{aR}$ as a polynomial of $aR$. For example, we can expand $e^{aR}$ up to its second order term as
\begin{equation}\label{2nd_expand}
    \Omega = e^{aR} \approx I + aR + a^2R^2/2
\end{equation}
to show that the optimal $aR$ is the solution of the following equation
\[ R(Q+Q^T) + (Q+Q^T)R = 2(Q^T - Q)/a \]
Here, we simply fix $R$ to the steepest ascent direction, i.e., $R=Q^T - Q$, and require $a>0$. Then, substituting (\ref{2nd_expand}) back to (\ref{max_tr}) and noting that ${\rm tr}(RQ) = {\rm tr}(Q^TQ) - {\rm tr}(Q^2)\ge 0$, we can determine the line search step size as 
\[
a = 
\begin{cases}
\mu_{\max} / \|R\|, & \text{if } {\rm tr}(R^2 Q) \ge 0 \\[6pt]
\min\left(-{\rm tr}(RQ)/{\rm tr}(R^2Q), \, \mu_{\max} /\|R\|\right), & \text{otherwise}
\end{cases}
\]
where $\mu_{\max}>0$ is a normalized maximum allowable step size to keep the truncation error under control. For example, by clamping the step size $a$ to $0.25/\|R\|$, the truncation error $\Omega^T\Omega - I = a^4R^4/4$ with expansion (\ref{2nd_expand}) is upper bounded as $\|\Omega^T\Omega - I\| = \left\| \left(0.25/\|R\|\right)^4  R^4/4 \right \| \le 1/4^5<0.001$. 

It is possible to expand $e^{aR}$ to terms of higher order for faster convergence. Note that the expansion to minimize $\|\Omega^T\Omega - I\|$ is generally not the same as the Taylor series of $e^{aR}$. For example, to expand $e^{aR}$ to terms up to the third order, we obtain 
\begin{equation}\label{3rd_expand}
    \Omega = e^{aR} \approx I + aR + a^2R^2/2 + a^3R^3/8
\end{equation}
and the truncation error is $I - \Omega^T\Omega = a^6R^6/64$. Noting that ${\rm tr}(RQ)\ge 0$ and ${\rm tr}(R^3Q)\le 0$, we can easily determine the line search step size for (\ref{3rd_expand}) as 
\[a=\min\left(\frac{ -{\rm tr}(R^2Q) - \sqrt{\left[{\rm tr}(R^2Q)\right]^2 - 1.5\,{\rm tr}(RQ){\rm tr}(R^3Q)} }{0.75\, {\rm tr}(R^3Q)}, \; \frac{\mu_{\max}}{\|R\|}\right)\]
when $Q^T\ne Q$ and thus ${\rm tr}(RQ)> 0$ and ${\rm tr}(R^3Q)< 0$. We typically set $\mu_{\max}=5/8$ to ensure that the truncation error is upper bounded by $(5/8)^6/64 < 0.001$. The fourth order expansion and truncation error are $\Omega = e^{aR} \approx I + aR + 0.5a^2R^2 + (2-\sqrt{2})a^3R^3/4 + (3-2\sqrt{2})a^4R^4/8$ and $\Omega^T\Omega - I\approx 0.00046a^8R^8$, respectively. The line search step size is the smaller positive root of a cubic equation clamped at the maximum value $1.1/\|R\|$ such that the truncation error $\|\Omega^T\Omega - I\|$ is upper bounded by $0.001$. However, we seldom go beyond the third order approximation in practice.  

A few notes on the effectiveness of the above rotations for imposing the SPD property. Clearly, any symmetric $Q$ is a stationary point of (\ref{opb}). Such symmetric $Q$ is a saddle point if it is not SPD. Let us only consider the stable stationary points, i.e., the maxima of (\ref{opb}). Note that the orthogonal group $O(n)$ is not connected. Hence, such rotations can only make $Q$ with a positive determinant SPD. For $Q$ with a negative determinant, all its eigenvalues, except for the one with the smallest absolute value, will be positive at the maximum. Hence, to make the rotations work, we should initialize $Q$ to a matrix in ${\rm GL}^+(n, \mathbb{R})$, say a SPD matrix. Still, in the domain of complex numbers, the unitary group $U(n)$ is always connected and such rotations can make any complex $Q$ SPD except for convergence to saddle points.       
\begin{figure}[h]
\centering
\includegraphics[width=0.7\textwidth]{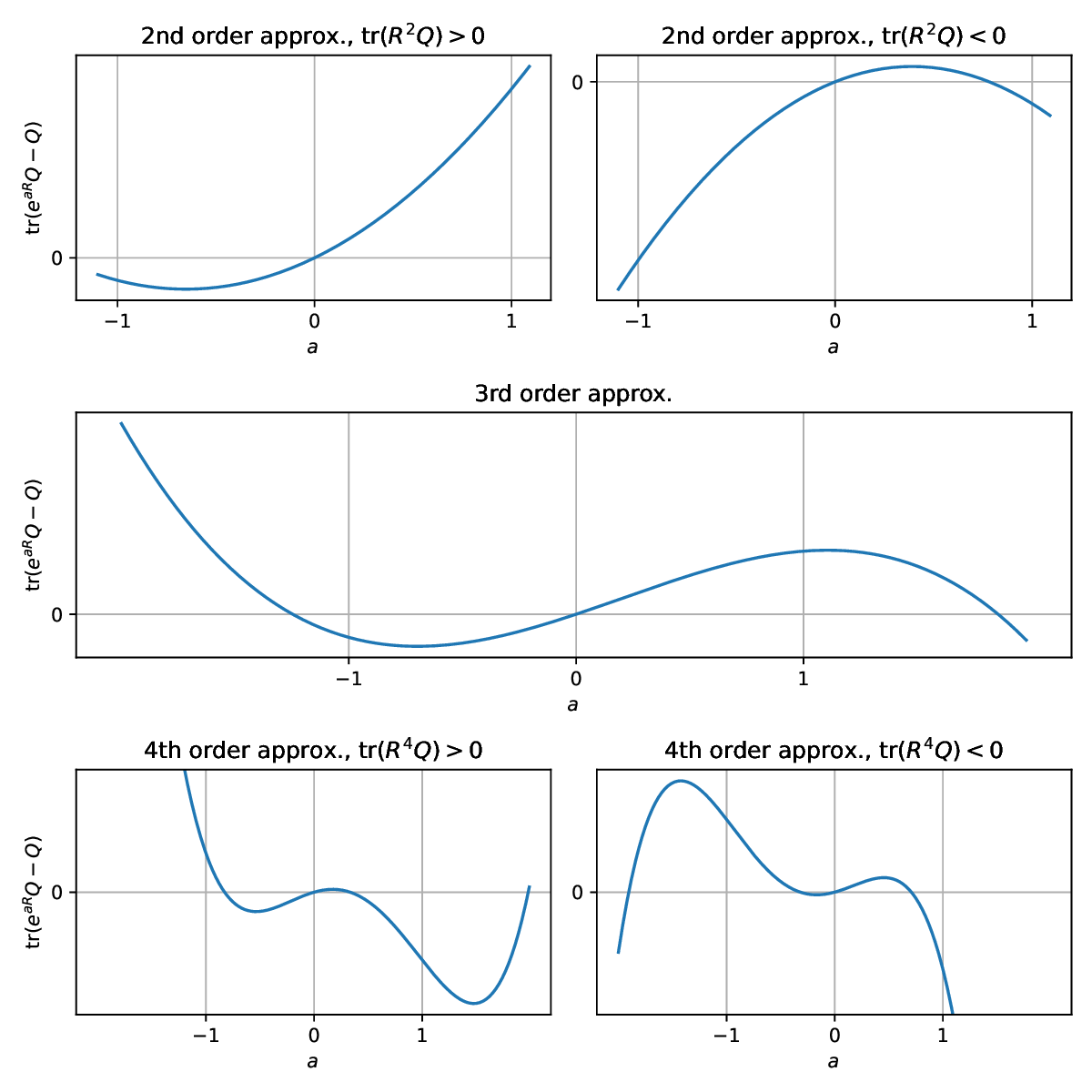}
\caption{Plots illustrating the possible shapes of $f(a)={\rm tr}(e^{aR}Q - Q)$ with different orders of approximation of $e^{aR}$, where $R=Q^T - Q$ is the rotation group generator. Note that ${\rm tr}(RQ)\ge 0$ and ${\rm tr}(R^3Q)\le 0$ and the equal sign holds only when $R=0$. }
\end{figure}

\subsubsection{A Quadratic Form for Updating $Q$} 

One simple method to ensure the SPD property is to update $Q$ with the following quadratic form,
\begin{equation}\label{update_quad2}
Q_{t+1} = \left[ I - \frac{\mu}{2L_t} (P_t h_th_t^T P_t - v_t v_t^T)\right] Q_t  \left[ I - \frac{\mu}{2L_t} (P_t h_th_t^T P_t - v_t v_t^T)\right] 
\end{equation}
In this way, we always have $Q_t\succ 0$ as long as $Q_0\succ 0$. Since both (\ref{update_QEQ}) and (\ref{update_quad1}) are stable with $Q_t\succ 0$, we see that (\ref{update_quad2}) is stable too as long as $\mu\ll 1$  such that the quadratic term $\mathcal{O}(\mu^2)$ is negligible. However, this quadratic term also prevents us from reformulating (\ref{update_quad2}) as a gradient descent method minimizing (\ref{P_criterion}) in any coordinate. To rigorously prove that (\ref{update_quad2}) converges for any symmetric $Q$, let us define the error signal $E=Phh^TP - vv^T$ and rewrite the update rule of $Q$ as
\begin{equation}\label{quad_deltaQ}
    \lim_{\mu\rightarrow 0}\delta Q = Q_{t+1} - Q_t \propto -( EQ + QE )
\end{equation}
On the other hand, the gradient of (\ref{P_criterion}) with respect to $Q$ is 
\begin{equation}\label{quad_gradientQ}
    \nabla_Q = 2(Qhh^T - Q^{-T}vv^T P^{-1})=2 Q^{-1} E P^{-1}
\end{equation}
Combining (\ref{quad_deltaQ}) and (\ref{quad_gradientQ}) and noting that $Q$, $E$ and $P$ all are symmetric, we have 
\begin{align*}
     {\rm tr}\left( (\delta Q)^T \; \nabla_Q  \right)  & \propto -{\rm tr}\left[  (EQ+QE) Q^{-1} E P^{-1}\right] \\
     & = -{\rm tr}(E^2P^{-1} + QEQ^{-1}EP^{-1}) \\
    &  = -{\rm tr}(EQ^{-1} Q^{-1} E + Q^{-1}E Q^{-1}E  ) \le 0
\end{align*}
where the last inequality is due to ${\rm tr}(A^TA + A^2) \ge 0$ with $A=Q^{-1} E$ here. Hence, (\ref{update_quad2}) is always stable for a small $\mu$ even if $Q$ loses its SPD property due to numerical round-off errors. 

\subsubsection{Connections to Newton-Schulz Iterations} 

It is not difficult to show that both (\ref{update_quad1}) and (\ref{update_quad2}) can be used to update $P$ directly by replacing $Q$ with $P$ \emph{if} we can ensure the SPD property of $P$. Basically, we define the local coordinate $dP = P^{0.5}\mathcal{E}P$, and take the gradient descent wrt $P'=\int P^{-0.5} (dP) P^{-1}$ to arrive at the following gradient descent method for minimizing (\ref{P_criterion})
\begin{equation}\label{update_quad3}
P_{t+1} = P_t - \frac{\mu}{L_t} (P_t h_th_t^T P_t - v_t v_t^T) P_t  
\end{equation}
After integrating out $v\sim \mathcal{N}(0, I)$ as $E_v[hh^T]=E_v[Hvv^TH]=H^2$ and $E_v[vv^T]=I$ from (\ref{update_quad3}), we can obtain the following simplified update rule for $P$,
\begin{equation}\nonumber  
P_{t+1} = \left(1 +  \frac{\mu}{L_t}\right)P_t - \frac{\mu}{L_t} P_t H^2 P_t^2   
\end{equation}
which is the same as the Newton-Schulz iteration (\ref{newton_P}) when $\mu/L_t = 0.5$. Similarly, (\ref{update_quad1}) can directly provide the following inverse fourth root Newton iteration method for a SPD matrix $A=H^2$,
\begin{equation}\label{4th_root}
   Q_{t+1} = Q_t - 0.25(P_tAP_t - I)Q_t 
\end{equation}
by letting $\mu=0.5$, $L=2$, $Q_0\propto I$ such that all the $Q_t$'s commute with $A$, and $\|P_0AP_0\| \le 1$ such that $L_0\le 2$. Since all the $Q_t$'s commute with $A$, they share the same eigenvectors as $A$. Thus, (\ref{4th_root}) essentially reduces to the inverse fourth root Newton iteration 
\[ q_{t+1} = 1.25 q_t - 0.25 a q_t^5  \]
for solving $q^{-4} - a=0$ with a scalar $a$ and starting from a positive initial guess $q_0$ that satisfies $aq_0^4\le 1$. However, the gradient descent method is more widely applicable and reliable than the Newton method, as it does not depend on the perfect knowledge of $H$, the commuting property between $Q$ and $H$, and the initial condition $\|P_0 H^2P_0\|\le 1$.  


Although it is possible to fit $P$ directly, there are a few caveats. With (\ref{update_quad3}), the following orthogonal Procrustes step cannot make $P$ perfectly symmetric, let alone its SPD property. Several orthogonal Procrustes steps might be required to make $P$ almost SPD. With (\ref{update_quad2}) and replacing $Q$ with $P$, $P$ could lose its SPD property due to numerical round-off errors, although it must be SPD in theory. It is common to force $P$ to be symmetric by $P\leftarrow (P+P^T)/2$. However, such an additive update could introduce zero or negative eigenvalues to $P$. Hence, either choice could be problematic in practice under certain circumstances, especially with half precision arithmetic.  

\subsection{Hessian Fitting with Local Coordinate $dQ=Q\mathcal{E}P$}

One last choice to derive a nontrivial inverse-free update rule for fitting $Q$ is to both left- and right-multiplying (\ref{gradient_glg}) with $QQ^T$ to have 
\begin{equation}\label{update_QEP}
Q_{t+1} = Q_t - \frac{\mu}{L_t}  Q_t(P_t h_th_t^T P_t - v_t v_t^T) Q_t^T Q_t 
\end{equation}
where similar to (\ref{Lt}), $L_t$ can be estimated as 
\begin{equation}\nonumber 
    \ell_t = \|Q_tP_th_t\|^2 + \|Q_t v_t\|^2, \quad L_t= \max\left(\beta L_{t-1} + (1-\beta)\ell_t, \; \ell_t\right) 
\end{equation}
On the other hand, the gradient of criterion (\ref{P_criterion}) wrt to $Q'=\int Q^{-1} (dQ) P^{-1}$ and the quadratic terms of $\mathcal{E}$ are
\begin{align*}
    & \nabla_{Q'} = 2(Phh^TP - vv^T) \\
    & h^TP\mathcal{E}^TP\mathcal{E}Ph + v^T \mathcal{E} P \mathcal{E} v + v^T \mathcal{E} P \mathcal{E}^T v + v^T \mathcal{E}^T P \mathcal{E}^T v
\end{align*}
respectively. 
Thus, (\ref{update_QEP}) is actually a gradient descent method for $Q$ minimizing criterion (\ref{P_criterion}) with the local coordinate defined as $dQ=Q\mathcal{E}P$. Furthermore, criterion (\ref{P_criterion}) is convex there with constraint $\mathcal{E}=\mathcal{E}^T$. 
However, it cannot be strongly convex there as the above quadratic terms can be arbitrarily small for $\|P\|\rightarrow 0$. Hence, this method could converge slowly if $Q$ starts from a poor initial guess. Nevertheless, empirical results suggest that this method is highly competitive for semi-quadratic problems where the curvature does not change dramatically. 

\subsection{Numerical Results}

We consider a gradient whitening problem to illustrate the properties of the above four inverse-free methods. The covariance matrix of the gradient is $H = {\rm hilb}(64) + 10^{-6}I$, where ${\rm hilb}(64)$ denotes the Hilbert matrix with size $64$. The condition number of $PHP$ indicates how well the gradients are whitened. Fig.~3 summarizes the comparison results. We see that (\ref{update_QEP}) converges the fastest, and (\ref{update_QEQ}) and (\ref{update_quad2}) tend to oscillate before convergence due to the difficulty in bounding the Hessians in their local coordinates. What is not shown here is that (\ref{update_QEP}) could converge slowly when starting from bad initial guesses. In practice, (\ref{update_quad1}) provides the best trade-off among performance, reliability, and complexity.   

\begin{figure}[h]
\centering
\includegraphics[width=0.5\textwidth]{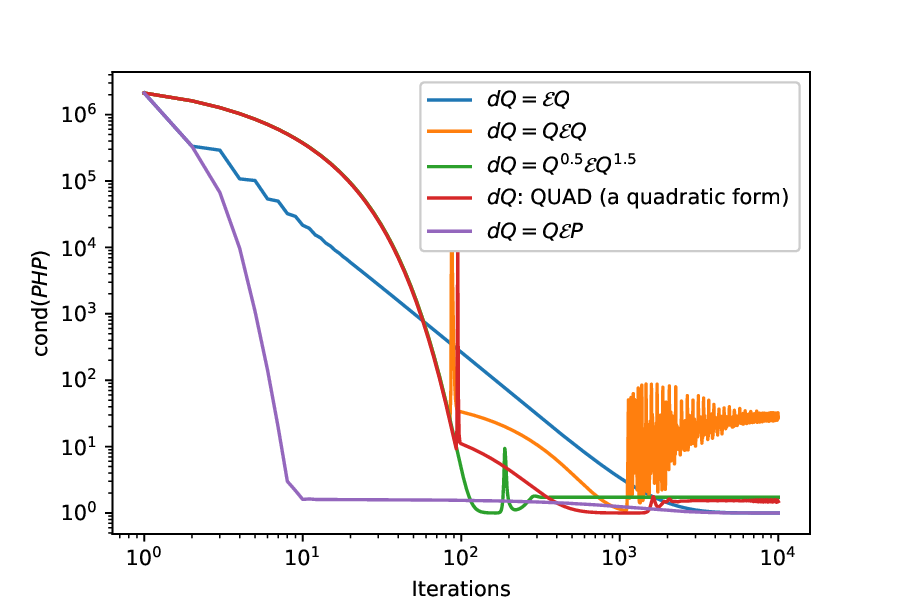}
\caption{The five Hessian fitting methods in (\ref{standard_method}), (\ref{update_QEQ}), (\ref{update_quad1}), (\ref{update_quad2}) and (\ref{update_QEP}) are compared on whitening gradient covariance matrix $H={\rm hilb}(64) + 10^{-6}I$. We start from $Q_0=I$, set $\beta=1$, and $\mu=1$ for (\ref{standard_method}) and (\ref{update_QEP}), and $\mu=0.1$ for (\ref{update_QEQ}) (\ref{update_quad1}), and (\ref{update_quad2}).   }
\end{figure}

\section{Sparse Lie Groups for Large Scale Hessian Fittings}

The above Hessian fitting methods are only suitable for small to medium-scale problems with roughly $n < 10^5$ parameters to optimize. For problems with millions to billions of parameters to learn, we need significantly sparser Hessian estimators. 

\subsection{Diagonal Preconditioner}

Let $Q={\rm diag}(q)$, where $q$ is a vector without zero  element. Note that $\|A\|_2=\max_{1\le i\le n} |a_{i,i}| $ when $A$ is diagonal. It is trivial to follow the same procedures leading to (\ref{standard_method}) to derive the following SGD method for updating $Q$ in the group of invertible diagonal matrices, 
\begin{equation}\label{method_diag}
    q_{t+1} = q_t - \frac{\mu}{L_t} \left[ \left(q_t \cdot h_t\right)^2  - \left(v_t \cdot /q_t\right)^2 \right] \cdot q_t 
\end{equation}
where the $\cdot$ suggests that the operations between two vectors are element-wise ones, and $\ell_t = \max\left(\left(q_t\cdot h_t\right)^2  + \left(v_t\cdot/q_t\right)^2 \right) $ is used to update $L_t$. It is also trivial to derive the inverse-free versions for updating $Q$ and we skip the details here.    

\subsection{Kronecker Product Preconditioner}

The Kronecker product preconditioner has been discussed in \cite{Li18} and \cite{psgd_affine}. Note that the KFAC \cite{KFAC} and Shampoo \cite{hess_kron, Shampoo1,Shampoo2,Shampoo3} like preconditioners are quite different from ours   even though they all share the same Kronecker product forms. The KFAC and Shampoo preconditioners can only approximate the inverse of Hessians even when they can be factorized into the assumed Kronecker product forms, while ours could converge to the exact solution given the correct factorization form.  

Without loss of generality, we only consider the case of $Q=Q_2\otimes Q_1$. The step index subscript $t$ is temporarily suppressed to simplify the writing.  A Kronecker preconditioner that has more than two factors has similar forms and can be best implemented with Einstein sums or tensor dot products. Interested readers can refer to our Pytorch code and supplementary materials \cite{psgd_kron}.   By Proposition 5, we choose local coordinate $dQ = (\mathcal{E}_2\otimes I + I \otimes \mathcal{E}_1) Q$. To obtain the partial derivative wrt $Q_2$, let us freeze $Q_1$ and only consider the update of $Q_2$. Thus, we have $dQ_2=\mathcal{E}_2 Q_2$ and, 
\begin{equation*}
dQ = \left(\mathcal{E}_2Q_2\right)\otimes Q_1 =\left(\mathcal{E}_2\otimes I\right)\left(Q_2\otimes Q_1\right)=\left(\mathcal{E}_2\otimes I\right) Q
\end{equation*}
Then, by replacing the $\mathcal{E}$ in (\ref{dcP_two_choices}) with $\left(\mathcal{E}_2\otimes I\right)$, we obtain the second order approximation for $c(P+dP)-c(P)$. Also, noting that $\|\mathcal{E}_2\otimes I\|_2=\|\mathcal{E}_2\|_2$, we can follow the procedures leading to   (\ref{triangular_update}) to bound the Hessian with respect to $\mathcal{E}_2$ to derive a step size normalized update rule for $Q_2$.  In the same way,  we can freeze $Q_2$ and have $dQ = \left(I \otimes \mathcal{E}_1 \right) Q$ by taking $dQ_1=\mathcal{E}_1 Q_1$.  Again, noting that $\|I \otimes \mathcal{E}_1\|_2=\|  \mathcal{E}_1\|_2$, we can derive another step size normalized update rule for $Q_1$. We skip these tedious details and directly provide the  SGD update for $Q$ as below,
\begin{align}\nonumber 
A_t & = Q_{1,t}{\rm uvec}(h_t)Q_{2,t}^T, \; B_t = Q_{2,t}^{-T} \left[ {\rm uvec}(v_t)\right]^T Q_{1,t}^{-1} \\ \nonumber 
Q_{1, t+1} & = \left[ I - \frac{\mu}{L_{1,t}}(A_tA_t^T-B_t^TB_t) \right]_R Q_{1,t} \\ \label{method_affine}
Q_{2, t+1} & = \left[ I - \frac{ \mu}{L_{2,t}}(A_t^TA_t-B_tB_t^T) \right]_R Q_{2,t} 
\end{align}  
where ${\rm uvec}(v_t)$ and ${\rm uvec}(h_t)$ recover vectors $v_t$ and $h_t$ to their original matrix forms, respectively, and $\ell_{1,t} =\|A_tA_t^T + B_t^TB_t\|_2 $ and $\ell_{2,t} =\|A_t^TA_t + B_tB_t^T\|_2 $ are used to update $L_{1,t}$ and $L_{2,t}$, respectively. They take the same forms as those given in \cite{Li18} and \cite{psgd_affine} except for the step size normalizers.  

We can follow the same procedures as shown in Section IV to derive the inverse-free Kronecker product preconditioner update rules. For example, with local coordinate $dQ = Q\mathcal{E} Q$, we have 
\begin{align}\nonumber 
dQ = (Q_2\otimes Q_1)(\mathcal{E}_2\otimes I + I \otimes \mathcal{E}_1) (Q_2\otimes Q_1)
\end{align}
Then, we take the derivative wrt $\mathcal{E}_1$ and $\mathcal{E}_2$ and bound their Hessians to arrive at the following update rules for $Q_1$ and $Q_2$,
\begin{align}\nonumber 
A_t & = Q_{1,t}^T Q_{1,t}{\rm uvec}(h_t)Q_{2,t}^TQ_{2,t}, \; B_t = \left[ {\rm uvec}(v_t)\right]^T  \\ \nonumber 
Q_{1, t+1} & =Q_{1,t}  - \frac{\mu}{L_{1,t}} Q_{1,t} (A_tA_t^T-B_t^TB_t)   \\  \nonumber
Q_{2, t+1} & = Q_{2,t} - \frac{ \mu}{L_{2,t}}Q_{2,t}(A_t^TA_t-B_tB_t^T)  
\end{align}  
We do not list all the inverse-free Kronecker product preconditioner update methods here since they all have similar forms to the ones given in Section IV.

\subsection{Low-Rank Approximation (LRA) Preconditioner}

Similar to the L-BFGS method,  we have developed a LRA preconditioner in \cite{li2022black}. Let us revisit it and refine its implementations. We can have the following two choices for the definition of $Q$,
\begin{align}\nonumber 
{\rm choice \,I: }\quad Q &= (I + UV^T){\rm diag}(d) \\ \label{UVd}
{\rm choice \, II: }\quad Q &= {\rm diag}(d) (I + UV^T) 
\end{align}
where $U$ and $V\in \mathbb{R}^{n\times r}$ with $0\le r\ll n$, and vector $d$ has no zero element such that  ${\rm diag}(d)$ is in the group of diagonal matrices. Both choices of (\ref{UVd}) have the same representation capacity for nonsingular matrices with form ${\rm diag}(d')+U'V'^T$. Still, the first choice is easier to work with when we update $Q$ as $dQ=\mathcal{E}Q$. 

The key to working with the LRA preconditioner is to notice the Lie group structures of the $Q$ in (\ref{UVd}). Clearly, ${\rm diag}(d)$ is already in the group of diagonal matrices. Noting that 
\begin{equation}\label{group_U}
  e^{\varepsilon G_1 V^T} =I +  \left( \sum_{i\ge 1} \frac{\varepsilon^i (G_1V^T)^{i-1}G_1}{i!} \right)V^T=I + \mathcal{E}_1V^T
\end{equation}
readers familiar with Lie groups can see that set $\{ I+UV^T | U\in \mathbb{R}^{n\times r} , \,\det(I+V^TU)\ne0,  V {\rm \, is \, fixed}\}$ forms a Lie group. Similarly, with a given $U$, set $ \{ I+UV^T | V\in \mathbb{R}^{n\times r},\,\det(I+V^TU)\ne0, U {\rm \, is\, fixed} \}$ forms another Lie group. We can exploit these Lie group structures to update $U$, $V$ and $d$ efficiently. For example, to update $U$, from (\ref{group_U}), we can let $dQ$ be
\begin{equation*}
dQ = e^{\varepsilon G_1 V^T} Q - Q = \mathcal{E}_1V^T Q
\end{equation*}
and then simply replace the $\mathcal{E}$ in (\ref{dcP_two_choices}) with $ \mathcal{E}_1V^T$ to obtain the second order approximation of $c(P+dP)-c(P)$. Similarly, to update $V$, we switch to the other group and  let $dQ$ be
\begin{equation*}
dQ = e^{\varepsilon U G_2^T} Q - Q = U \mathcal{E}_2^T Q
\end{equation*}
We omit the lengthy derivations here and only present the final update rules for $d$, $U$ and $V$  as below,
\begin{align}\nonumber 
a_t =&  Q_t h_t, \quad b_t = Q_t^{-T}v_t \\ \nonumber
d_{t+1} = & \left[1 - \frac{\mu}{L_{t,d}}\left(h_t\cdot (P_th_t) - v_t\cdot (P_t^{-1}v_t)\right)\right] \cdot d_t  \\ \nonumber 
U_{t+1} = & U_t - \frac{\mu}{L_{t, u}}  (a_ta_t^T - b_tb_t^T)V_t(I+V_t^TU_t)  \\ \label{method_lra}
V_{t+1} = & V_t -  \frac{\mu}{L_{t,v}}  (I+V_tU_t^T) (a_ta_t^T - b_tb_t^T)U_t 
\end{align} 
where the $\cdot$ suggests that operations between two vectors are element-wise ones, $\ell_{t,d}=\max|h_t\cdot (P_th_t)| + \max|v_t\cdot (P_t^{-1}v_t)|$, $\ell_{t,u} = \|a_t\| \|V_tV_t^Ta_t\| + \|b_t\|\|V_tV_t^Tb_t\|$ and $\ell_{t,v} = \|a_t\| \|U_tU_t^Ta_t\| + \|b_t\|\|U_tU_t^Tb_t\| $ are used to update $L_{t, d}$, $L_{t, u}$ and $L_{t,v}$, respectively.  
Interested readers can refer to the supplementary materials of \cite{li2022black} for further details.  The only difference is that in (\ref{method_lra}), we have normalized the step sizes with information from the second order derivatives. Note that within a single step, we can only update either $U$ or $V$, not both simultaneously.  

A few notes before concluding the discussions on the LRA preconditioner. The matrix inverse in (\ref{method_lra}) can be cheaply calculated with the Woodbury matrix identity \cite{Golub} since typically $0\le r \ll n$. We usually update either $U$ or $V$, not both, within a single update step as set $ \{ I+UV^T | U, V\in \mathbb{R}^{n\times r}\,{\rm and}\, \det(I+V^TU)\ne 0 \}$ no longer forms a Lie group. Unfortunately, this also prevents us from developing any inverse-free LRA preconditioner. Our LRA preconditioner is quite different from the L-BFGS method. The L-BFGS method is essentially a limited number of steps of expansion of the BFGS iterations and has no memory of Hessian-vector products beyond $r$ steps. Clearly, (\ref{method_lra}) is not a simple expansion of (\ref{standard_method}) up to $r$ steps. It keeps track of the best approximation of $H^{-1}$ with the LRA forms in (\ref{UVd}). Lastly, (\ref{UVd}) does not define a unique representation for the LRA preconditioner since $UV^T = (UX^{-T})(VX)^T$ for any invertible matrix $X$. This could cause numerical issues. One way to resolve this ambiguity is to find the representation with the minimum $\|U\|_F^2 + \|V\|_F^2$ by solving the following problem,
\begin{equation}\label{UV_balance}
    \min_X \|UX^{-T}\|_F^2 + \|VX\|_F^2
\end{equation}
The exact solution for (\ref{UV_balance}) requires three eigenvalue decompositions (EVDs). Here, we just do a simple gradient descent based online adjustment for $U$ and $V$ as below,
\begin{align*}
    E = & \frac{U^TU - V^TV}{{\rm tr}(U^TU + V^TV)} \\
    U^{\rm new} = & U\left( I - \mu E + 0.5\mu^2 E^2 \right) \\
    V^{\rm new} = & V \left( I + \mu E + 0.5 \mu^2 E^2 \right)
\end{align*}
where $E$ gives a normalized mismatch error matrix between $U$ and $V$, $0<\mu<1$ is the gradient descent step size, and a quadratic term of $E$ is used to bound the adjustment error as 
\[ \|U^{\rm new} (V^{\rm new})^T - UV^T\| = \|0.25 \mu^4 U E^4 V^T\| \le 0.25 \mu^4 \|UV^T\| \]
We typically set $0<\mu\le 0.25$ to ensure at least three decimal digits of relative accuracy. 

\subsection{Summary of the Lie Group Hessian Fitting Methods and Numerical Results}

Table III summarizes the  information of the above five basic types of  Lie group preconditioners. With the diagonal preconditioner as the baseline and assuming that $\theta={\rm vec}(\Theta)$ and $\Theta\in \mathbb{R}^{m\times m}$, we see that the two dense preconditioner fitting methods have quadratic or higher complexities. The affine and LRA preconditioners are two promising choices. Between these two preconditioners, the  affine one requires more computations, while the LRA one with $r\ll m$ needs more memory space. It is possible to mix these basic building blocks to derive richer forms of Lie group preconditioners, e.g., the scaling-normalization preconditioner in \cite{psgd_affine} which only has a sublinear space complexity, i.e., $\mathcal{O}(m)$, although its computational complexity still is $\mathcal{O}(m^2)$.  The inverse-free versions have similar complexities, but they tend to have better numerical properties. 
 
\begin{table}[h]
\caption{Lie groups for Hessian fittings. Assuming that $\theta={\rm vec}(\Theta)$ with $\Theta\in \mathbb{R}^{m\times m}$ for the storage and computation numbers. }
\begin{center}
\begin{tabular}{ |c|c|c|c| } 
 \hline
 $Q$ & $Q_{t+1}\leftarrow Q_t$ & Storage  & Computation \\ 
 \hline 
 GL$(n, \mathbb{R})$ & (\ref{standard_method}) & $\mathcal{O}(m^4)$ & $\mathcal{O}(m^4)$ \\ 
 Triangular & (\ref{triangular_update})  & $\mathcal{O}(m^4)$ & $\mathcal{O}(m^6)$ \\
 Diagonal & (\ref{method_diag}) &  $\mathcal{O}(m^2)$  &  $\mathcal{O}(m^2)$  \\
 $Q_2\otimes Q_1$ & (\ref{method_affine}) & $\mathcal{O}(m^2)$ &  $\mathcal{O}(m^3)$    \\
 $(I+UV^T){\rm diag}(d)$ & (\ref{method_lra}) & $\mathcal{O}(rm^2)$ & $\mathcal{O}(rm^2)$   \\
 \hline
\end{tabular}
\end{center}
\end{table}

Next, we present a few numerical results to show the effectiveness of the proposed preconditioners. With $v\sim \mathcal{N}(0, I)$ and imposing constraint $Q_0\propto I$, we can easily derive the optimal initial guess for $Q$ as $Q = \left(h_1^Th_1/n\right)^{-1/4}I$ given the first sample $h_1\in\mathbb{R}^{n\times 1}$ by minimizing (\ref{P_criterion}). When $Q$ has the direct sum form $Q_t=\oplus_i Q_{i,\,t}$, we choose the minimum scale to initialize all $Q_{i,\,0}$ such that $Q_0\propto I$.   

\subsubsection{Numerical Example I: Tensor Rank Decomposition}

The results shown in Fig.~2 clearly suggest the potential of PSGD for traditional convex or mathematical optimizations \cite{Boyd}. What is interesting is that PSGD seems to be less vulnerable to saddle points than those off-the-shelf optimizers since it does not rely on line search. Let us consider a  tensor rank decomposition problem defined by
\begin{equation}\label{trd}
\min_{x, y, z}\sum_{i=1}^{I}\sum_{j=1}^{J}\sum_{k=1}^{K} \left( \tau_{i,j,k}- \sum_{r=1}^{R} x_{ri} y_{rj}z_{rk} \right)^2
\end{equation}
where $\tau$ is a given tensor, and parameters $x, y$ and $z$ are to be solved. The loss landscape is riddled with saddle points, e.g., $x=y=0$ and an arbitrary $z$. This type of problem has a few specialized optimizers like the alternating least squares (ALS) methods. Here, we only consider the general purpose optimizers. Fig.~4 shows a set of  typical convergence curves.  The stairway-like convergence curves of GD and L-BFGS clearly indicate their difficulties in escaping  saddle points as the gradients around there are too small. PSGD works equally well for positive and non-positive definite Hessian matrices and is less vulnerable to saddle points. All three versions of PSGD work well here.       

\begin{figure}[h]
\centering
\includegraphics[width=0.5\textwidth]{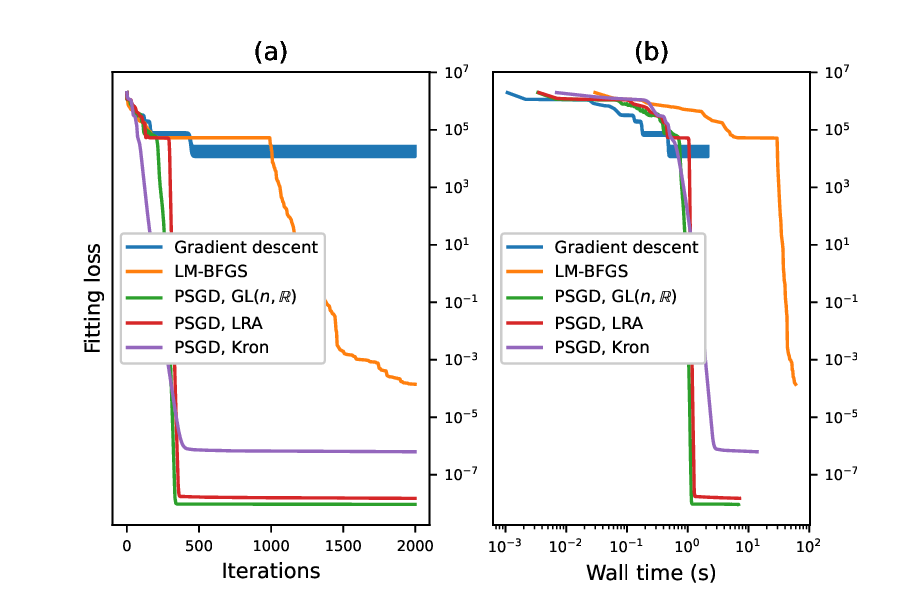}
\caption{Comparisons on  the tensor rank decomposition problem of (\ref{trd}), where  $R=10, I=20, J=50$, $K=100$, $\tau_{i,j,k}= \sum_{r=1}^{R} a_{ri} b_{rj}c_{rk}$, and all the elements of $a$, $b$ and $c$  are drawn from $\mathcal{N}(0, 1)$.  Except for the PSGD LRA optimizer, the inverse-free PSGD with local coordinate $dQ=Q^{0.5}\mathcal{E}Q^{1.5}$    is used.      }
\end{figure}

Nevertheless, line search is an important step in convex optimization to accelerate convergence. For example, let us consider the strongly convex loss $f(\theta)=\theta^2 - 4\sqrt{\theta}$ defined in $\theta\in\mathbb{R}^{+}$. Starting from the initial guess $\theta=0$,  Newton's method will get stuck there since $\lim_{\theta\rightarrow 0} \frac{df(\theta)}{d\theta}/\frac{d^2f(\theta)}{d\theta^2}=-2\lim_{\theta\rightarrow 0}\theta=0$.  GD and line search together help to push $\theta$ to the basin of attraction; only there does Newton's method converge quadratically. The purpose of the above experiment is to show that PSGD can operate without line search. This could help to enhance the traditional BFGS like quasi-Newton optimizers by relaxing their line search conditions.   

\subsubsection{Numerical Example II: Transformer Model Trainings}

Please check our Pytorch demos for detailed experimental setups. 

\emph{Vision Transformer (ViT) Model}: This example considers a small ViT model for the CIFAR10 image recognition task. We have matched Adam's and PSGD's hyper-parameter settings so that their only difference is the preconditioner. Fig.~5 shows a set of typical training loss and test accuracy results. Clearly, all five PSGD Kronecker product preconditioners have significantly outperformed the Adam optimizers.   

\begin{figure}[h]
\centering
\includegraphics[width=0.8\textwidth]{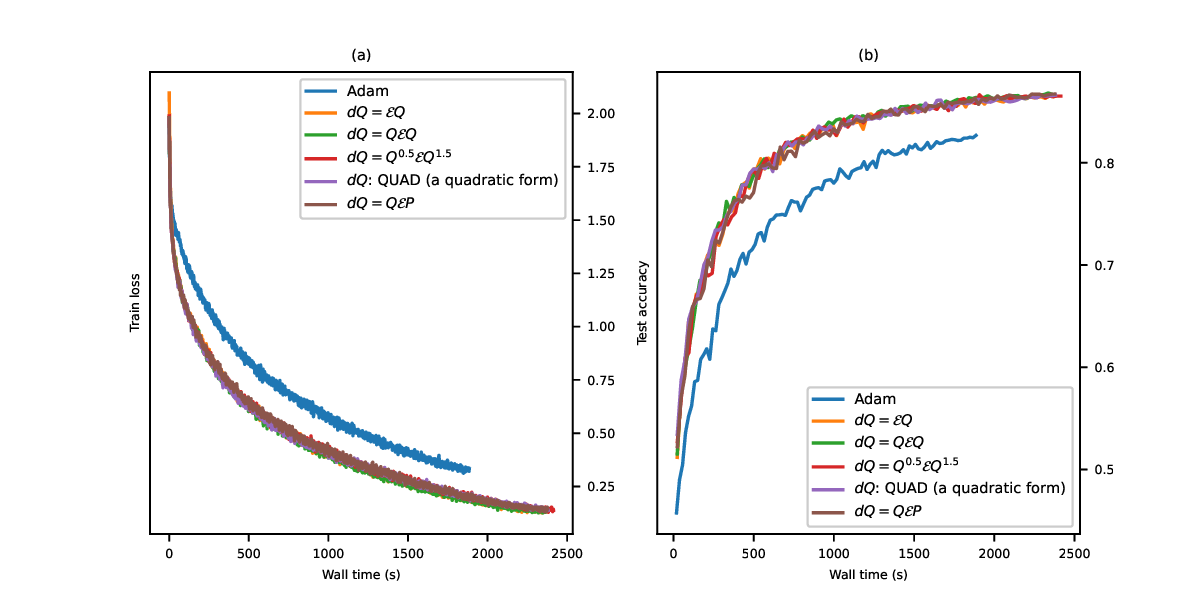}
\caption{ Comparisons between the  PSGD gradient whitening preconditions and Adam on a small ViT model for the CIFAR10 image recognition task.  PSGD and Adam use the same parameter learning rate, $10^{-3}$. They mainly differ by their used preconditioners.  }
\end{figure}

\emph{Generative Pre-trained Transformer (GPT) Model}: This example considers a small GPT2 model for the WiKi text dataset next token prediction task. We have trained the models exclusively with half precision. Preconditioner for PSGD is fitted with local coordinate $dQ=Q^{0.5}\mathcal{E} Q^{1.5}$. The other four choices for $dQ$ are not considered due to the high cost of each run. For PSGD, we choose to whiten the momentum, and its learning rate is roughly $\sqrt{(1+0.9)/(1-0.9)}\approx 4$ times smaller than that of Adam. Please refer to (\ref{fisher_hess}) for the rationale for reducing the learning rate when the momentum is whitened. Fig.~6 shows a set of typical training and validation losses. Again, we see that PSGD has significantly outperformed Adam.   

\begin{figure}[h]
\centering
\includegraphics[width=0.8\textwidth]{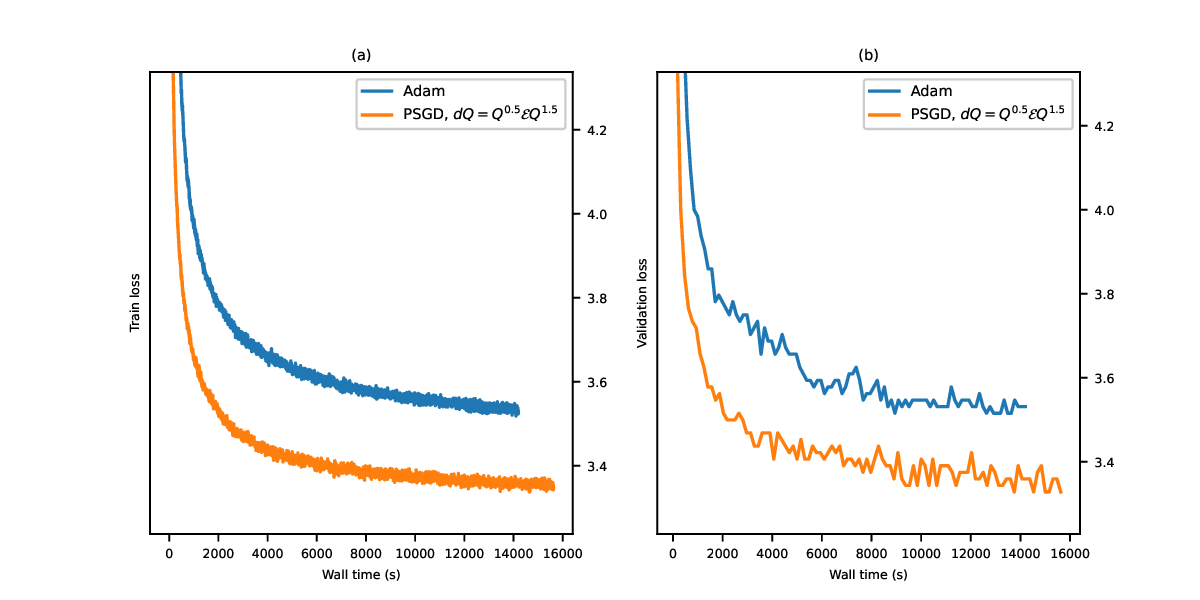}
\caption{ Comparisons between the PSGD Kronecker momentum whitening and Adam optimizers on a small GPT2 model training task with half precision. Adam uses learning rate $10^{-3}$, and PSGD uses $10^{-3}/4$ as it whitens the momentum (see (\ref{fisher_hess}) for explanation).   }
\end{figure}

\section{Conclusion}

In this report, we have studied a wide range of stochastic Hessian fitting methods using stochastic gradients or Hessian-vector products as their inputs.  Exact closed-form solutions are available only for a few simple cases such as a diagonal or a dense matrix preconditioner form. They do converge sublinearly to the optimal solution, but may involve numerically challenging matrix operations like inverses and eigenvalue decompositions. The stochastic Hessian fitting criterion from the preconditioned stochastic gradient descent (PSGD) method is thoroughly analyzed in the Euclidean space, the manifold of symmetric positive definite (SPD) matrices and a variety of Lie groups. It is shown to be strongly convex under mild conditions in the Lie group GL$(n, \mathbb{R})$ with one form of local coordinate definition, see Proposition 5 in Section III.A for details. This unique discovery facilitates our designs of several Lie group stochastic Hessian fitting methods with linear convergences, self-normalized step sizes, and different time and space complexities, see Table III in Section V.D for a summary. We further have proposed four inverse-free Hessian fitting methods by choosing different local coordinates for $Q$. All of them have similar forms and can be analyzed similarly.   Practically, these Lie group stochastic Hessian fitting methods successfully avoid any numerically problematic operations like large matrix inverses or decompositions. Numerical results have confirmed our theoretical analyses, and demonstrated their efficiency and numerical reliability for real world stochastic optimization problems. 


%

\section*{Acknowledgment}

I am deeply grateful to those pioneer users of PSGD, especially Omead Pooladzandi and Evan Walters, who have shared with me their valuable feedback and insights to help improve the theory of PSGD.  

\section*{Appendix I: Hessian Fitting in the Euclidean Space and Manifold of SPD Matrices}

\subsection{Hessian Fitting in the Euclidean Space}

\subsubsection{Gradient Descent}

An obvious choice is to update $P$ with the gradient given in (\ref{cP_gradient}). Let us start from an initial guess  $P_0\succ 0$, and update $P$ at step $t$ with SGD as 
\begin{equation}\label{cP_hv_gd}
    P_{t+1} = P_t - \mu(h_th_t^T - {P}_t^{-1} v_t v_t^T {P}_t^{-1})
\end{equation}
where $\mu$ is a small enough positive step size, $(v_t, h_t)=(v_t, Hv_t)$ with $v_t\sim \mathcal{N}(0, I)$ is a pair of vector and its associated Hessian-vector product at step $t$. Note that $P_t$ is always symmetric as long as $P_0$ is. 

From the proof of Proposition 1, we see that the Hessian of $c(P)$ with respect to $P$ is neither lower nor upper tightly bounded. Thus, it is generally not easy to characterize the convergence rate of the iteration in (\ref{cP_hv_gd}). Still, following the regret analysis \cite{adagrad, Adam}, we can show that with a proper constant step size $\mu$, the sequence given by (\ref{cP_hv_gd}) converges sublinearly to $H^{-1}$ when $H\succ 0$. We only sketch the key steps here. Assume that $\mu$ is small enough so that we always have $P_t\succ 0 $ as long as $P_0\succ 0$. Then, via the convexity of $c(P;v,h)$, we can show that  
\begin{align*}
& \quad c(P_t;v_t, h_t) - c(H^{-1};v_t, h_t) \\
&  \le \frac{\left\| P_t - H^{-1}\right\|_F^2 - \left\| P_{t+1} - H^{-1}\right\|_F^2}{2\mu} + \frac{\mu}{2} \left\|  \frac{\partial c(P_t;v_t,h_t)}{\partial P} \right\|_F^2
\end{align*}
Then, we take expectation on both sides of the above equation with respect to $v$, and sum over the time index $t$ to have  
$$c\left(\sum_{i=1}^t P_i/t\right) - c(H^{-1}) = \mathcal{O}(1/\sqrt{t})$$
by setting $\mu\propto 1/\left( \sqrt{t} \, \max_{1\le i\le t} E_{v_i} \left[ \left\|  \frac{\partial c(P_i; v_i, h_i)}{\partial P} \right\|_F \right] \right) $. Still, the regret analysis does not exclude possibly better convergence results of (\ref{cP_hv_gd}) with time-varying step sizes. 

\subsubsection{Newton's Method}

Another choice is to take the expectation on the left side of (\ref{opt_dP}) to have 
\begin{equation}\label{newton_P_no_commute}
    H^2 - P^{-2} + P^{-2} dP\, P^{-1} + P^{-1} dP\, P^{-2}=0
\end{equation}
and then solve for the optimal $dP$ either directly as
\begin{align}\nonumber 
   &  \:\: {\rm vec}(dP) =  \left( P^{-1}\otimes P^{-2} + P^{-2}\otimes P^{-1} \right)^{-1} {\rm vec}\left(P^{-2} - H^2\right) \\ \nonumber 
   & =  \left( I \otimes P^{-1} + P^{-1}\otimes I  \right)^{-1} (P\otimes P) {\rm vec}\left(P^{-2} - H^2\right) \\ \label{newton_stable_one}
   &  =  \left( I \otimes P^{-1} + P^{-1}\otimes I  \right)^{-1}  {\rm vec}\left(I - P H^2 P\right)
\end{align}
or via the matrix sign function iteratively \cite{mtxsignfunction}, which is clear after rewriting (\ref{newton_P_no_commute}) into the following familiar continuous Lyapunov function form,
\[ P^{-1} dP + dP \, P^{-1} = I - PH^2P \] 
where $H^2$ is given or estimated from $H^2=E[hh^T]$ by replacing  $E[\cdot]$ with sample averages. 
Eq. (\ref{newton_stable_one}) or the method in \cite{mtxsignfunction} gives a numerically stable but costly solution for $dP$. 

But, if we assume that $P$ and $dP$ commute such that $P^{-2} dP\, P^{-1} = P^{-1} dP\, P^{-2}$, then (\ref{newton_P_no_commute}) immediately gives the optimal $dP$ as below,  
\begin{equation}\label{dP_no_vec}
    dP = -0.5 P (H^2 - P^{-2}) P^2 = 0.5 (P - PH^2P^2)
\end{equation}
Now, (\ref{dP_no_vec}) suggests the following Newton's method for updating $P$,
\begin{equation}\label{newton_P}
    P_{t+1} = 1.5P_t  -0.5 P_t H^2P_t^2 
\end{equation}
which basically reduces to the well-known Newton-Schulz iteration for calculating the square root of $H^2$. 

\emph{Proposition 2: }Starting from an initial guess $P_0\succ 0$ that commutes with $H$ and satisfies condition $0<\lambda(HP) < (\sqrt{17}-1)/2$, Newton's method  (\ref{newton_P}) converges to $H^{-1}$ quadratically.  

\emph{Proof: }First, let us show that  $P_{t+1}$ commutes with $H$ if $P_t$ does. With (\ref{newton_P}), we have 
\begin{align*}
    & P_{t+1}H - H P_{t+1}  \\
    = & 1.5(P_tH - HP_t) - 0.5(P_tH^2P_t^2H - H P_tH^2P_t^2) \\
    = & -0.5(P_tH^3P_t^2 - P_tH^3P_t^2) = 0
\end{align*}
Hence, $P_t$ always commutes with $H$ as long as $P_0$ does. Also, we can easily show that $P_t$ is symmetric as long as $P_0$ is.  

Now, let us define $R_t = HP_t - I$. Similarly, we can show that $P_t$, $H$ and $R_t$ all commute with each other and are symmetric. Then, by (\ref{newton_P}), we have 
\begin{align}\nonumber 
    R_{t+1} = & 1.5 HP_t  - 0.5 HP_t H^2 P_t^2 - I \\ \nonumber
     = & 1.5 HP_t - 0.5 (HP_t)^3 - I \\ \nonumber
     = & 1.5 (R_t + I) - 0.5(R_t + I)^3 - I \\ \label{R_newton_recursive}
     = & -0.5 (R_t + 3I) R_t^2 
\end{align}
We require 
\[ \|0.5 (R_t + 3I)R_t \|_2 < 1 \]
to make (\ref{R_newton_recursive})  a contraction mapping such that $\|R_{t+1}\|_2 < \|R_t\|_2$,  which suggests that the eigenvalues of $P_tH$ should be bounded by
\begin{equation}\label{PH_eig_bounds}
    0<\lambda (P_tH) < \frac{\sqrt{17} - 1}{2}
\end{equation}
Then, with the bounds given in (\ref{PH_eig_bounds}) and the last line of (\ref{R_newton_recursive}), we can show the following  quadratic convergence,
\[ \frac{\|R_{t+1}\|_2}{\|R_t\|_2^2} = 0.5 \|R_t + 3I\|_2 < \frac{\sqrt{17}+3}{4} \]
Note that the lower bound of $\lambda (P_tH)$ in (\ref{PH_eig_bounds}) already implies that $H$ cannot be singular such that $H^{-1}$ exists. Also, please note that for the rate of quadratic convergence, it is sufficient to show that ${\|R_{t+1}\|_2}/{\|R_t\|_2^2}$ is tightly upper bounded, not necessarily strictly upper bounded by $1$. \hfill $\square$

It is trivial to choose a starting point $P_0$ that commutes with $H$, e.g., a $P_0\propto I$. Indeed, starting from a $P_0\propto I$, we see that $P_t$ is a linear combination of the terms in $\{I, H^2, H^8, \ldots \}$ and thus commutes with $H$. However, in practice, the Newton's method given in (\ref{newton_P}) is vulnerable to round-off errors that eventually break the commuting property between $P_t$ and $H$. 

Another closely related method is the modified Newton-Schulz iteration \cite{NSiteration}, which is numerically stable and more commonly used to calculate $(H^2)^{-0.5}$ iteratively. We seldom use these Newton's methods in practice. We present them here mainly for theoretical interest, especially their connections to the iterative methods for matrix square root calculations \cite{mtxsignfunction, NSiteration}. 

\subsubsection{A Few Closed-Form Solutions}

For the sake of completeness, let us also briefly discuss the convergence rate of a few closed-form solutions in the Euclidean space. The most straightforward and commonly used choice is  
$$P_t=\left(\left(P_0^{-2}+\sum_{i=1}^t h_ih_i^T\right)/(t+1)\right)^{-0.5}$$ 
e.g.,  methods from the AdaGrad family \cite{adagrad}, which can be put in a  fancier form as below
\begin{equation}\label{closed-form-solutionI}
    P_{t+1} = \left(\frac{t+1}{t+2}P_t^{-2} + \frac{1}{t+2}h_th_t^T \right)^{-0.5}
\end{equation}
Since $h_t=Hv_t$ and $v_t\sim\mathcal{N}(0,I)$, we have 
\begin{align*}
    P_t = & \left( H \left(P_0^{-2}/(t+1) + \sum_{i=1}^t v_iv_i^T/(t+1) \right) H \right)^{-0.5} \\
    = & \left( H \left( I + \mathcal{O}(1/t) \right) H \right)^{-0.5} \\
    = & H^{-1} + \mathcal{O}(1/t)
\end{align*}
Hence, the closed-form solution given by (\ref{closed-form-solutionI}) converges to $H^{-1}$ sublinearly with rate $\mathcal{O}(1/t)$ when $H\succ 0$. 

Another closed-formed solution can be obtained from (\ref{opt_dP}) by solving for $P$ directly after setting $dP=0$. Specifically, (\ref{opt_dP}) suggests that a stationary point of $P$ satisfies 
\begin{equation}\label{closed-form-solutionII}
   P \left( \sum_{i=1}^t h_ih_i^T \right) P-  \sum_{i=1}^t v_iv_i^T  = 0
\end{equation}
Then, we can directly solve the above Riccati equation for $P$  when both $\sum_{i=1}^t v_iv_i^T$ and $\sum_{i=1}^t h_ih_i^T$ are not singular, see the proof of Proposition 3 in \cite{Li18}. Clearly, (\ref{closed-form-solutionII}) reduces to (\ref{closed-form-solutionI}) for $t\rightarrow\infty$. But this solution is exact even with a finite $t$ as long as the solution is unique. Similar to  Newton's methods, this closed-form solution is mainly for theoretical interest. 

One more  family of solutions worth  mentioning here are the BFGS like methods that modify $P_t$  with certain rank-2 updates such that $P_{t+1} h_t = v_t$ \cite{Boyd}. They are essentially iterative methods for Hessian fittings, but with a few heuristic closed-form update rules for $P$. These methods have two major limitations: the pair $(v_t, h_t)$ needs to satisfy the curvature condition $v_t^Th_t>0$; they diverge easily with noisy Hessian-vector products as inputs. Note that generally, these heuristic methods could diverge or might not converge to $\left(H^2\right)^{-0.5}$.

\subsection{Hessian Fitting in the Manifold of SPD}

One possible way to define a local coordinate for $P$ in the manifold of SPD is to let $dP$ be
\begin{equation}\label{local_coordinate_spd}
    dP = P \mathcal{E}  + \mathcal{E} P^T
\end{equation}
where $\mathcal{E}$ is a sufficiently small matrix, but not necessarily symmetric in general. Clearly, $dP$ is symmetric if $\mathcal{E}$ is, regardless of the symmetry property of $P$. On the other hand, we can show that $\mathcal{E}$ is symmetric if $P\succ 0$. 

\emph{Proposition 3: }Eq.~(\ref{local_coordinate_spd}) induces a new local coordinate system for the manifold $\{P|P\succ 0\}$.

\emph{Proof:} We are to show that (\ref{local_coordinate_spd}) defines a bijective mapping between $dP$ and $\mathcal{E}$ when $P\succ 0$, which is partially done after we vectorize both sides of (\ref{local_coordinate_spd}) to connect $dP$ to $ \mathcal{E} $ as,
\begin{equation}\label{vec_E_1}
    {\rm vec}(\mathcal{E}) = (I \otimes P + P\otimes I )^{-1} {\rm vec}(dP) 
\end{equation}
Still, we need to show that (\ref{vec_E_1}) preserves the symmetry properties of $dP$ and $\mathcal{E}$. Clearly, a symmetric $\mathcal{E}$ maps to a symmetric $dP$ by (\ref{local_coordinate_spd}). The left part is to show that (\ref{vec_E_1}) also maps a symmetric $dP$ to a symmetric $\mathcal{E}$. To begin with, let us introduce a  permutation matrix $\Pi$ such that ${\rm vec}(A^T) = \Pi \,{\rm vec}(A) $. Then, for any matrix $X$ with the same size as $P$, we have
\begin{align*}
    (P\otimes I + I\otimes P)\Pi \, {\rm vec}(X) = & (P\otimes I + I\otimes P){\rm vec}(X^T) \\
    = & {\rm vec}(X^TP + PX^T ) \\
    = & {\rm vec}\left( (PX + XP)^T \right) \\
    = & \Pi {\rm vec}(PX+XP) \\
    = & \Pi (I\otimes P + P\otimes I) {\rm vec}(X)
\end{align*}
Since $X$ is arbitrary, the above equation suggests that 
\begin{equation}\label{PIIP+permute_Pi}
    \Pi^{-1} (P\otimes I + I\otimes P) =  (I\otimes P + P\otimes I) \Pi^{-1}
\end{equation}
Then, from (\ref{vec_E_1}), we have
\begin{align*}
    {\rm vec}(\mathcal{E}^T) = & \Pi\, {\rm vec}(\mathcal{E}) \\
    = & \Pi (I \otimes P + P\otimes I )^{-1} {\rm vec}(dP) \\
    = &  \left( (I \otimes P + P\otimes I )\Pi^{-1} \right)^{-1} {\rm vec}(dP) \\
    = & \left(\Pi^{-1}  (I \otimes P + P\otimes I )\right)^{-1} {\rm vec}(dP) \\
    = &  (I \otimes P + P\otimes I )^{-1} \Pi {\rm vec} (dP) \\
    = &  (I \otimes P + P\otimes I )^{-1} {\rm vec} (dP^T) \\
    = & (I \otimes P + P\otimes I )^{-1} {\rm vec} (dP) \\
    = & {\rm vec}(\mathcal{E}) 
\end{align*}
where we have used (\ref{PIIP+permute_Pi}) from the third to the fourth line. Thus, we have  $\mathcal{E}=\mathcal{E}^T$ when $dP=dP^T$. Finally, note that the condition $P\succ 0$ not only implies $dP=dP^T$, but also makes (\ref{vec_E_1}) bijective. This completes the proof.  \hfill $\square$

By Proposition 3, (\ref{local_coordinate_spd}) induces a new coordinate system for the manifold of SPD matrices.  For a positive scalar $P$, integration on the right side of (\ref{vec_E_1}) gives  the closed-form solution for transforming  $P$ to the new one,  i.e., $P' = 0.5\log P + {\rm constant}$. In general, we have no need to write down this transform explicitly for a matrix $P$, and (\ref{local_coordinate_spd}) is enough for our purpose. 

Now, we can rewrite the $dc(P;v,h)$ in (\ref{dcP_2nd}) in the new coordinate as below after ignoring term $ \mathcal{O}(\mathcal{E}^3)$,
\begin{align}\nonumber 
   &  \quad dc(P; v, h) \\ \nonumber 
    & =  h^T (P\mathcal{E} + \mathcal{E}P^T) h  - v^T P^{-1}  (P\mathcal{E} + \mathcal{E}P^T) P^{-1} v  +  v^T P^{-1} (P\mathcal{E} + \mathcal{E}P^T) P^{-1} (P\mathcal{E} + \mathcal{E}P^T) P^{-1} v   \\ \label{dcP_2nd_E} 
    & =  h^T P \mathcal{E}h + h^T\mathcal{E}Ph - v^T\mathcal{E}P^{-1}v - v^TP^{-1}\mathcal{E}v +  v^T\mathcal{E}^2 P^{-1} v  + v^T \mathcal{E}P^{-1}\mathcal{E}v + v^T P^{-1}\mathcal{E}P\mathcal{E}P^{-1}v + v^TP^{-1}\mathcal{E}^2v 
\end{align}
where in the last line we have assumed that $P=P^T$. 
Again, (\ref{dcP_2nd_E}) is a quadratic  function of $\mathcal{E}$, and we can solve for the $\mathcal{E}$ that reduces $dc(P; v, h)$ the most from the following equation,
\begin{align} \nonumber 
    & \quad P hh^T + hh^T P- vv^TP^{-1} - P^{-1}vv^T + vv^TP^{-1}\mathcal{E}^T   + \mathcal{E}^T vv^TP^{-1}  + vv^T\mathcal{E}^T P^{-1} + P^{-1} \mathcal{E}^Tvv^T  \\   \label{opt_E_SPD_manifold} 
    &  + P^{-1}vv^T P^{-1}\mathcal{E}^TP + P\mathcal{E}^TP^{-1}vv^TP^{-1}  + P^{-1}vv^T\mathcal{E}^T + \mathcal{E}^TP^{-1} vv^T = 0 
\end{align} 
Then, we can identify the gradient in the new coordinate from (\ref{opt_E_SPD_manifold}),  and update $P$ with SGD given pair $(v_t, h_t)$ at step $t$ as,
\begin{align} \nonumber 
    \mathcal{E}_t = & -\mu( P_th_th_t^T + h_th_t^TP_t - v_tv_t^T P_t^{-1} - P_t^{-1} v_t v_t^T ) \\ \label{GD_on_SPD_manifold}
    P_{t+1} = & P_t + ( P_t \mathcal{E}_t + \mathcal{E}_t P_t )
\end{align}
Clearly, with (\ref{GD_on_SPD_manifold}), $P_t$ is always symmetric as long as $P_0$ is. 

\emph{Proposition 4: }Starting from an initial guess $P_0\succ 0$ that commutes with $H$ and with a constant step size $0<\mu< 0.5/\max_t {\lambda_{\max}(H+H^2P_t)}$, the sequence $P_t$ given by (\ref{GD_on_SPD_manifold}) converges to $H^{-1}$ linearly with rate $1 - 8\mu \, \lambda_{\min}(H) $. 

\emph{Proof: }We take expectation on both sides of (\ref{GD_on_SPD_manifold}) to have 
\begin{align}\nonumber 
    & \quad E[P_{t+1}] \\ \nonumber 
    & =  E[P_t] - \mu \left( E[P_t^2] H^2 + 2E[P_t H^2 P_t] + H^2E[P_t^2] - 4I \right) \\   \label{with_E1} 
    & =  E[P_t] - \mu \left( \left(E[P_t]\right)^2H^2 + 2 E[P_t]H^2E[P_t]   + H^2 \left(E[P_t]\right)^2 - 4I + \mathcal{O}( \left(P_t - E[P_t]\right)^2 )\right)  
\end{align}
where $E[\cdot]$ is a shorthand for $E_{v_1\sim \mathcal{N}(0, I), \ldots, v_t\sim \mathcal{N}(0, I)}[\cdot]$. 
The term $\mathcal{O}( \left(P_t - E[P_t]\right)^2 )$ in the last line of (\ref{with_E1}) is negligible with a small $\mu$. We further drop the notation $E[\cdot]$ in (\ref{with_E1}), and just use $P_t$ to denote its expectation to simplify our writing.     

First, we are to show that $P_{t+1}$ commutes with $H$ if $P_t$ does. Assuming $P_tH = HP_t$, we can simplify (\ref{with_E1}) to
\begin{equation}\label{P_t1_proposition3}
    P_{t+1} = P_t - 4\mu (H^2 P_t^2 - I)
\end{equation}
Then, it is not difficult to show that $P_{t+1}$ also commutes with $H$, similar to the proofs in Proposition 2. 
Thus, $P_t$ always commutes with $H$ as long as $P_0$ does. 

Second, let us define fitting error $R_t = HP_t - I$ and show that $\|R_t\|_2\rightarrow 0$ for  $t\rightarrow\infty$. Clearly, $H$, $P_t$ and $R_t$ all are symmetric and commute with each other. From (\ref{P_t1_proposition3}), we have  
\begin{align}\nonumber 
    R_{t+1} = & H P_{t+1}  - I \\ \nonumber 
    = & H P_t - 4\mu H\left((HP_t)^2 - I \right) - I \\ \nonumber 
    = & (R_t + I) - 4\mu H \left( (R_t + I)^2 - I \right) - I \\ \nonumber 
    = & (I - 8\mu H - 4\mu H R_t) R_t \\ \label{Rt_to_Rt1}
    = & (I - 4\mu H -4\mu H^2P_t) R_t 
\end{align}
From the last line of (\ref{Rt_to_Rt1}), we see that $\mu$ needs to satisfy 
\[ 0 < \mu   < \frac{1}{2\lambda_{\max}(H + H^2P_t)} \]
to ensure the monotonic convergence of $P_t$. Noting that $\|R_t\|_2\rightarrow 0$ for $t\rightarrow\infty$ and $[H, R_t]=0$, the second to the last line of (\ref{Rt_to_Rt1}) gives  the following asymptotic convergence rate, 
\[ 1 > \frac{\|R_{t+1}\|_2}{\|R_t\|_2} = \|I - 8\mu H\|_2 \ge 1 - 8\mu\, \lambda_{\rm min}(H)   \]
Thus, $P_t$ converges linearly to $H^{-1}$ when $\lambda_{\rm min}(H)>0$. \hfill $\square$

Again, similar to the proof of Proposition 2, the condition that $P$ and $H$ commute simplifies the proof, but could be too strict to be satisfied in practice. Yet, the linear convergence rate $1 - 8\mu \, \lambda_{\min}(H) $ matches well with the observations. 

Compared with SGD in the Euclidean space, linear convergence of the iterations given by (\ref{GD_on_SPD_manifold}) with a small enough constant step size is a good sign of progress. Intuitively,  from (\ref{local_coordinate_spd}), we see that $dP$ is roughly proportional to $P$. Hence, (\ref{local_coordinate_spd}) mimics a Euclidean space varying step size adaptation.      
Following a similar way for the proof of Proposition 1, from (\ref{dcP_2nd_E}), we can show that the Hessian of $c(P)$ in the new coordinates is 
\[ S^T\left[ 3(P^{-1}\otimes I) + P^{-2}\otimes P \right] S  \]
which again is neither lower nor upper tightly bounded. Thus, the criterion (\ref{P_criterion}) still is not strongly convex.  We have only taken a small step forward. 

It is interesting to check Newton's methods in the new coordinates as well. Following the developments of the Newton's methods in the Euclidean space, we can take expectation on the left side of (\ref{opt_E_SPD_manifold}) and solve for the optimal $\mathcal{E}$ directly. Generally, the two sets of Newton's methods are different as the coordinate transform defined by (\ref{local_coordinate_spd}) is not linear. However, if we assume that $\mathcal{E}$ and $P$ commute, we can readily show that $dP$ and $P$ commute as well, and eventually the two Newton's methods are the same in this case. 

\section*{Appendix II: Implementation Details}
\setcounter{subsection}{0}

We have provided a basic Pytorch PSGD implementation at \url{https://github.com/lixilinx/psgd_torch/blob/master/psgd.py}.   

\subsection{Matrix Spectral Norm Estimation}

The spectral norm is used in the Lipschitz constant estimation of the PSGD preconditioner fitting criterion and solving the online orthogonal Procrustes problem. Its estimation is crucial to the preconditioner update rule of PSGD. 

\subsubsection{A Few Deterministic Bounds} 

There are abundant matrix norm bounds. But we prefer some very computationally cheap and still tight enough ones. Given a matrix $A\in \mathbb{R}^{n\times n}$, let us introduce notations 
\begin{align*}
    \|A\|_{\max}  & = {\max}_{1\le i, j\le n} |a_{ij}| \\
    \|A\|_1 & = {\max}_{1\le j\le n} \sum_{i=1}^n |a_{ij}| \\
    \|A\|_\infty & = {\max}_{1\le i\le n} \sum_{j=1}^n |a_{ij}| \\
    \alpha & = \sqrt{{\max}\left( {\max}_{1\le i\le n} \sum_{j=1}^n a_{ij}^2, \, \max_{1\le j\le n} \sum_{i=1}^n a_{ij}^2 \right)}
\end{align*}
and give the following bounds for $\|A\|_2$ without proof, 
\begin{align*}
    \frac{1}{\sqrt{r}}\|A\|_F & \le \|A\|_2 \le \|A\|_F \\
    \|A\|_{\max} & \le \|A\|_2 \le n \|A\|_{\max} \\
    \frac{1}{\sqrt{n}}\|A\|_1 & \le \|A\|_2 \le \sqrt{n}\|A\|_1 \\
    \frac{1}{\sqrt{n}}\|A\|_\infty & \le \|A\|_2 \le \sqrt{n}\|A\|_\infty \\
    \alpha & \le \|A\|_2 \le \sqrt{n}\alpha
\end{align*}
where $r$ is the rank of $A$. When $r\ll n$, $\|A\|_F$ gives a tight upper bound for $\|A\|_2$. In general, $\alpha$ gives the tightest lower bound for $\|A\|_2$. Still, we can further tighten it by a few power iterations.  Without loss of generality, we assume that $\alpha$ equals the vector norm of the $i$-th column  of $A$, i.e., $a_i$. Then, with only two matrix-vector products, a significantly tighter lower bound is given by 
\begin{equation}\label{power_iter}
    \|A\|_2 \ge \frac{\|AA^T a_i\|}{\|A^T a_i\|}
\end{equation}

\subsubsection{Subspace Iteration for Spectral Norm Estimation}

The power iteration method in (\ref{power_iter}) cannot take full advantage of parallel processing provided by today's hardware. For our purpose, we only need a rough spectral norm estimation. Subspace collapse is not a concern here. Thus, we have skipped the subspace orthogonalization step. The procedures are as below. We first draw a random matrix $V\in \mathbb{R}^{n\times k}$, where $k$ is the dimension of the subspace. Then we rotate $V$ with the Householder transformation such that its centroid, i.e., the mean of its columns, aligns with the $a_i$ in (\ref{power_iter}). Lastly, we do the power iteration for each column of $V$ independently without the orthogonalization step, and the maximum spectral norm estimation is used as the final answer. Since we only perform very few steps of iterations, the alignment step is crucial for the robustness of this method. In our implementations, we choose subspace dim $k=32$ and take four steps of subspace iteration without the orthogonalization step.   

\subsection{Regularization for Preconditioner Fitting}

Let us revisit the preconditioner fitting criterion (\ref{P_criterion}). Clearly, when $v\sim\mathcal{N}(0, I)$, the closed-form solution for $P$ is $P=(E[hh^T])^{-0.5}$ for Hessian fitting or $P=(E[gg^T])^{-0.5}$ for gradient whitening. The matrix inverse here could be problematic when $E[hh^T]$ or $E[gg^T]$ is almost singular. To address this issue, one can modify criterion (\ref{P_criterion}) to 
\begin{equation} \label{damping}
    (h + \nu)^T P (h + \nu) + v^T P^{-1} v
\end{equation}
where $\nu \sim \mathcal{N}(0, \eta^2  I)$ is the damping noise independent of $v$. We can integrate out $\nu$ to simplify (\ref{damping}) to 
\begin{equation} \label{damping_integrated_out}
    h P h + v^T P^{-1} v + \eta^2 {\rm tr}(P)
\end{equation}
Now, it is clear from (\ref{damping_integrated_out}) that the damping noise regularizes the preconditioner estimation such that $P \preceq I/\eta$ when $v\sim\mathcal{N}(0, I)$. Either (\ref{damping}) or (\ref{damping_integrated_out}) can be used to regularize the fitting of $P$ and they lead to similar additional complexity. Practically, since $\eta\ll 1$, say $\eta=10^{-9}$ in our implementation, (\ref{damping_integrated_out}) requires higher precision than (\ref{damping}). Even with (\ref{damping}), the damping noise could still be dropped out due to the lack of precision. To avoid this situation, we recommend sampling $\nu$ from $\nu \sim \mathcal{N}\left(0, \eta^2  I + {\rm eps}^2 \cdot {\rm diag}(|h|^2)\right)$, where ${\rm eps}$ is the machine precision. In this way, $E[(h + \nu)(h + \nu)^T]$ cannot be singular.  
 
\subsection{Negative Log-likelihood Losses and Gradient/Momentum Whitening Preconditioner}

\subsubsection{Relationship between Hessian and FIM}

Most machine learning optimization problems exclusively deal with the minimization of negative log-likelihood (NLL) losses. Let $f(x;\theta)=\log p(x;\theta)$ be a differentiable logarithmic probability density function (pdf) of the random variable $x$ parameterized by $\theta$. Due to the relationship,
\[ E\left[ \left(\frac{\partial  f(x;\theta) }{\partial \theta }\right) \left(\frac{\partial f(x;\theta) }{\partial \theta }\right)^T \right] = - E\left[  \frac{\partial^2 f(x;\theta) }{\partial 
\theta^T \partial \theta } \right] \]
we see that the Hessian, Fisher information matrix (FIM), and empirical FIM matrix are all the same when $z$ indeed follows the assumed form of distribution. Specifically, let $g$ be the gradient averaged over $B$ i.i.d. samples, and $m_t$ be the momentum updated as 
\[ m_{t+1} = \beta m_t + (1-\beta)g_t \]
where $0\le \beta<1$. Then, we have 
\begin{equation}\label{fisher_hess}
    {\rm Hessian} = {\rm FIM} = B\,E_z[gg^T] = B \frac{1+\beta}{1-\beta} E_z[mm^T]
\end{equation}
Hence, it is well motivated to use the gradient or momentum whitening preconditioner for NLL minimization. 
Note that the four terms in (\ref{fisher_hess}) are not necessarily the same when $z$ deviates from the assumed form of distribution. 

Let us consider a simple example to clarify the above statements. Let $z_i$, $1\le i\le T$, be the given data points. We are to fit their distribution as a normal one with unknown $\mu$ and $\sigma^2$. Hence, the parameters are $\theta=(\mu, \sigma)$. The NLL is given by 
$$ E_z\left[ (z-\mu)^2/(2\sigma^2) + \log\sigma \right] $$ 
where $E_z[\cdot]$ denotes the sample average over $z$. The per-sample gradient and Hessian are given by 
\begin{align*}
    g_z = &  \begin{bmatrix}
    (\mu-z)/\sigma^2  \\
    1/\sigma - (\mu-z)^2/\sigma^3 
\end{bmatrix} \\ 
    H = & E_z \begin{bmatrix}
    1/\sigma^2 &  2(z-\mu)/\sigma^3  \\
    2(z-\mu)/\sigma^3 & 3 (\mu-z)^2/\sigma^4 - 1/\sigma^2   
\end{bmatrix} 
\end{align*}
respectively. The FIM is ${\rm diag}(1/\sigma^2, 2/\sigma^2)$, which only depends on the assumed pdf, i.e., $\mathcal{N}(\mu, \sigma^2)$, not on the actual distribution of $z$. The empirical Fisher, $E_z[g_z g_z^T]$, is not necessarily diagonal in general. The Hessian is even not necessarily positive definite, say for $\sigma\rightarrow\infty$. Nevertheless, they all reduce to the same form when $z\sim\mathcal{N}(\mu, \sigma^2)$. Under the weaker condition of $E_z[g_z]=0$, we  see that $H$ still reduces to the FIM, but the empirical FIM can be different from the true FIM if $z$ is not Gaussian. In general, we can show that $H$ and FIM are the same when $E_z[g_z]=0$ and $p(\cdot; \theta)$ is from the exponential family. Further discussions of these topics are beyond the scope of this report. 

\subsubsection{On the Auxiliary Variable $v$} 

One more topic is on whether we should integrate out $v$ for updating the gradient or momentum whitening preconditioner as it appears to be a nuisance variable in (\ref{P_criterion}). Actually, the term $v^TPv$ can be regarded as a Hutchinson trace estimator, since its expectation is ${\rm tr}(P)$. Noting that $E_{v\sim\mathcal{N}(0,I)}[Q^{-T}vv^TQ^{-1}] = Q^{-T}Q^{-1}$, 
(\ref{standard_method}) becomes 
\begin{equation}\label{integrate_out_v}
    Q_{t+1} = Q_t - \mu \frac{Q_tg_tg_t^TQ_t^T - Q_t^{-T}Q_t^{-1} }{\|Q_tg_t\|^2 + {\rm tr}(Q_t^{-T}Q_t^{-1}) }   Q_t
\end{equation}
after integrating out $v$ and replacing $h$ with $g$. Since $Q\in\mathbb{R}^{n\times n}$, (\ref{standard_method}) only has computational complexity $\mathcal{O}(n^2)$ while (\ref{integrate_out_v}) has complexity $\mathcal{O}(n^3)$. In general, we prefer to keep $v$ as an auxiliary variable if doing so can simplify the implementation and computations. Otherwise, we simply integrate it out. The same argument holds for the comparison between (\ref{damping}) and (\ref{damping_integrated_out}). Integrating out $\nu$ does not necessarily simplify the computations there.  

\subsection{On the LRA Groups}

Readers not familiar with Lie groups can easily verify that set $\{ I+UV^T | U\in \mathbb{R}^{n\times r} , \,\det(I+V^TU)\ne0,  V {\rm \, is \, fixed}\}$ along with the matrix multiplication operation forms a Lie group by showing that it includes the identity matrix (when $U=0$), each of its elements has an inverse when $\det(I+V^TU)\ne0$, and lastly, matrix multiplication is associative. Similarly, with a given $U$, set $ \{ I+UV^T | V\in \mathbb{R}^{n\times r},\,\det(I+V^TU)\ne0, U {\rm \, is\, fixed} \}$ forms another Lie group. However, generally, set $ \{ I+UV^T | U, V\in \mathbb{R}^{n\times r}\,{\rm and}\, \det(I+V^TU)\ne 0 \}$ does not form a group since the products of its two elements may no longer belong to the same set. Hence, we choose to update either $U$ or $V$, not both, at each step in our implementations. Please check the supplementary materials of \cite{li2022black} for further details.


\vfill

\end{document}